\documentclass[dvipsnames]{article} %
\usepackage{colm2024_conference}

\usepackage{booktabs}
\usepackage{graphicx}
\usepackage{enumitem}
\usepackage{wrapfig}
\usepackage{algorithm}
\usepackage{algpseudocode}
\usepackage[utf8]{inputenc}

\usepackage{xltabular}
\usepackage{adjustbox}
\usepackage{amsmath}
\usepackage{colortbl}
\usepackage[utf8]{inputenc}
\definecolor{lightgray}{rgb}{0.9,0.9,0.9}
\usepackage{caption}
\usepackage{subcaption}
\usepackage[svgnames]{xcolor}
\usepackage{setspace}
\usepackage{url}
\usepackage{multirow}
\usepackage{colortbl}
\usepackage{tabularx}
\usepackage{blindtext}
\usepackage{pgfplots}
\pgfplotsset{compat=1.18} 
\usepackage{tikz}
\usetikzlibrary{er,positioning,bayesnet}
\usepackage{makecell}
\usepackage{tipa}
\usepackage{siunitx}
\usepackage{nicefrac}
\usepackage{tocloft}
\usepackage{listings}
\usepackage{xltabular}
\usepackage{adjustbox}
\usepackage{xurl}
\usepackage{rotating}
\usepackage[normalem]{ulem}
\usepackage{multicol}
\usepackage{titlesec}
\usepackage[most]{tcolorbox}
\usepackage{transparent}

\usepackage{amsthm}

\usetikzlibrary{shadows}

\renewcommand{\sectionautorefname}{Section}
\renewcommand{\subsectionautorefname}{Section}

\makeatletter
\g@addto@macro\appendix{%
  \def\subsectionautorefname{Appendix}%
  \def\sectionautorefname{Appendix}%
}
\makeatother

\newcommand{\blue}[1]{\textcolor{blue}{#1}}
\newcommand{\green}[1]{\textcolor{ForestGreen}{#1}}
\newcommand{\purple}[1]{\textcolor{purple}{#1}}

\newtcolorbox{codeblock}{
  colback=gray!10,   
  colframe=gray!50,  
  boxrule=0.5mm,     
  arc=2mm,           
  left=5pt,          
  top=5pt,          
  bottom=5pt,        
}

\newtcbtheorem{proposition}{Proposition}{
  fonttitle=\bfseries,
  fontupper=\normalfont,
  boxrule=0.5pt,
  left=3mm,      % 增加一点左边距
  right=4mm,
  top=2mm,
  bottom=2mm,
  clip upper=false,  % 默认是 false，但显式关闭更安全
  breakable=false,
  before upper=\setlength{\parindent}{1em}
}{prop}

\newtcbtheorem{corollary}{Corollary}{
  enhanced,
  colback=cyan!5!white,           % 背景色：极淡青
  colframe=cyan!45!blue!60,       % 边框色：青蓝混合
  fonttitle=\bfseries,            % 标题加粗
  fontupper=\normalfont,          % 正文正常字体（不斜体）
  arc=4mm,                        % �� 关键：整体圆角半径
  boxrule=1pt,
  drop shadow southwest,
  attach boxed title to top left={xshift=6mm, yshift=-2mm},
  boxed title style={
    colback=white,
    size=small,
    arc=2mm,                      % 标题块也圆角
    boxrule=0.8pt,
    left=2mm, right=2mm
  },
  before skip=10pt,
  after skip=10pt,
  left=5mm,
  right=5mm,
  top=3mm,
  bottom=3mm
}{cor} % 标签前缀，用于 \label{cor:xxx}

\useunder{\uline}{\ul}{}

\usepackage{amsmath,amsfonts,bm}

\def\eqref#1{equation~\ref{#1}}
\def\1{\bm{1}}

\DeclareMathAlphabet{\mathsfit}{\encodingdefault}{\sfdefault}{m}{sl}
\SetMathAlphabet{\mathsfit}{bold}{\encodingdefault}{\sfdefault}{bx}{n}

\renewcommand{\ghlink}{https://github.com/alibaba/ROLL}

\newcommand*\justify{%
  \fontdimen2\font=0.4em% interword space
  \fontdimen3\font=0.2em% interword stretch
  \fontdimen4\font=0.1em% interword shrink
  \fontdimen7\font=0.1em% extra space
  \hyphenchar\font=`\-% allowing hyphenation
}

\renewcommand{\texttt}[1]{%
  \begingroup
  \ttfamily
  \begingroup\lccode`~=`/\lowercase{\endgroup\def~}{/\discretionary{}{}{}}%
  \begingroup\lccode`~=`[\lowercase{\endgroup\def~}{[\discretionary{}{}{}}%
  \begingroup\lccode`~=`.\lowercase{\endgroup\def~}{.\discretionary{}{}{}}%
  \catcode`/=\active\catcode`[=\active\catcode`.=\active
  \justify\scantokens{#1\noexpand}%
  \endgroup
}

\definecolor{blueviolet}{RGB}{138,43,226}
\newtcolorbox{insightblock}{
  colback=blueviolet!5,   % 背景颜色，淡紫色
  colframe=blueviolet!50!black!50!,    % 边框颜色
  boxrule=0.5mm,       % 边框粗细
  arc=2mm,             % 边角弧度
  left=0pt,            % 左边距
  right=8pt,           % 右边距
  top=8pt,             % 上边距
  bottom=8pt,          % 下边距
}

\newtcolorbox{abstractbox}{
    colback=blue!5!white,     % 背景颜色（浅蓝色）
    frame empty,              % 边框颜色
    boxrule=1pt,              % 边框粗细
    arc=4mm,                  % 圆角
    left=8pt,                 % 左边距
    right=8pt,                % 右边距
    top=8pt,                  % 上边距
    bottom=8pt,                % 下边距
    opacityback=0.9
}

\newcommand{\SysName}{\textcolor{violet}{\texttt{ROLL Flash}}}

\title{Part \MakeUppercase{\romannumeral 2}: \SysName{} -- Accelerating RLVR and Agentic Training \\ with Asynchrony}
\author{
\textbf{Han Lu$^{12*}$, Zichen Liu$^{1*\dagger}$, Shaopan Xiong$^{1*\dagger}$,
Yancheng He$^{1*}$, Wei Gao$^{3*}$, \\
Yanan Wu$^{*,1}$,
Weixun Wang$^{1*\dagger}$,
Jiashun Liu$^{13}$,
Yang Li$^{12}$, 
Haizhou Zhao$^{1}$, \\
Ju Huang$^{1}$,
Siran Yang$^{1}$,
Xiaoyang Li$^{1}$,
Yijia Luo$^{1}$,
Zihe Liu$^{1}$, 
Ling Pan$^{3}$, \\
Junchi Yan$^{2}$, 
Wei Wang$^{3}$,
Wenbo Su$^{1}$,
Jiamang Wang$^{1}$,
Lin Qu$^{1}$,
Bo Zheng$^{1}$}
\\
\mbox{\small $^{1}$Alibaba Group, $^{2}$Shanghai Jiaotong University, $^{3}$Hong Kong University of Science and Technology} \\
}

\begin{document}

\vspace*{-25pt}
\maketitle

\begin{abstractbox}
\begin{center}
\textbf{\Large Abstract}
\end{center}
Synchronous Reinforcement Learning (RL) post-training has emerged as a crucial step for enhancing Large Language Models (LLM)  with diverse capabilities. However, many systems designed to accelerate RL post-training still suffer from low resource utilization and limited scalability. 
We present \SysName{}, a system that extends ROLL with native support for \emph{asynchronous} RL post-training. \SysName{} is built upon two core design principles: \emph{fine-grained parallelism} and \emph{rollout--train decoupling}. Guided by these principles, \SysName{} provides flexible programming interfaces that enable a fully asynchronous training architecture and support efficient rollout mechanisms, including queue scheduling and environment-level asynchronous execution.
Through comprehensive theoretical analysis and extensive experiments, we demonstrate that \SysName{} significantly improves resource utilization and scalability over synchronous RL post-training. \SysName{} achieves up to $2.24\times$ speedup on RLVR tasks, and $2.72\times$ on agentic tasks, using the same GPU budget as synchronous baselines. Furthermore, we implement several popular off-policy algorithms and verify that asynchronous training can achieve performance on par with synchronous training.
\end{abstractbox}

\begin{figure}[htbp]
    \vspace{-5pt}
    \centering

    % 左侧子图 (a)
    \begin{subfigure}[t!]{0.53\textwidth}
        \centering
        \includegraphics[width=\linewidth, height=0.5\textheight, keepaspectratio]{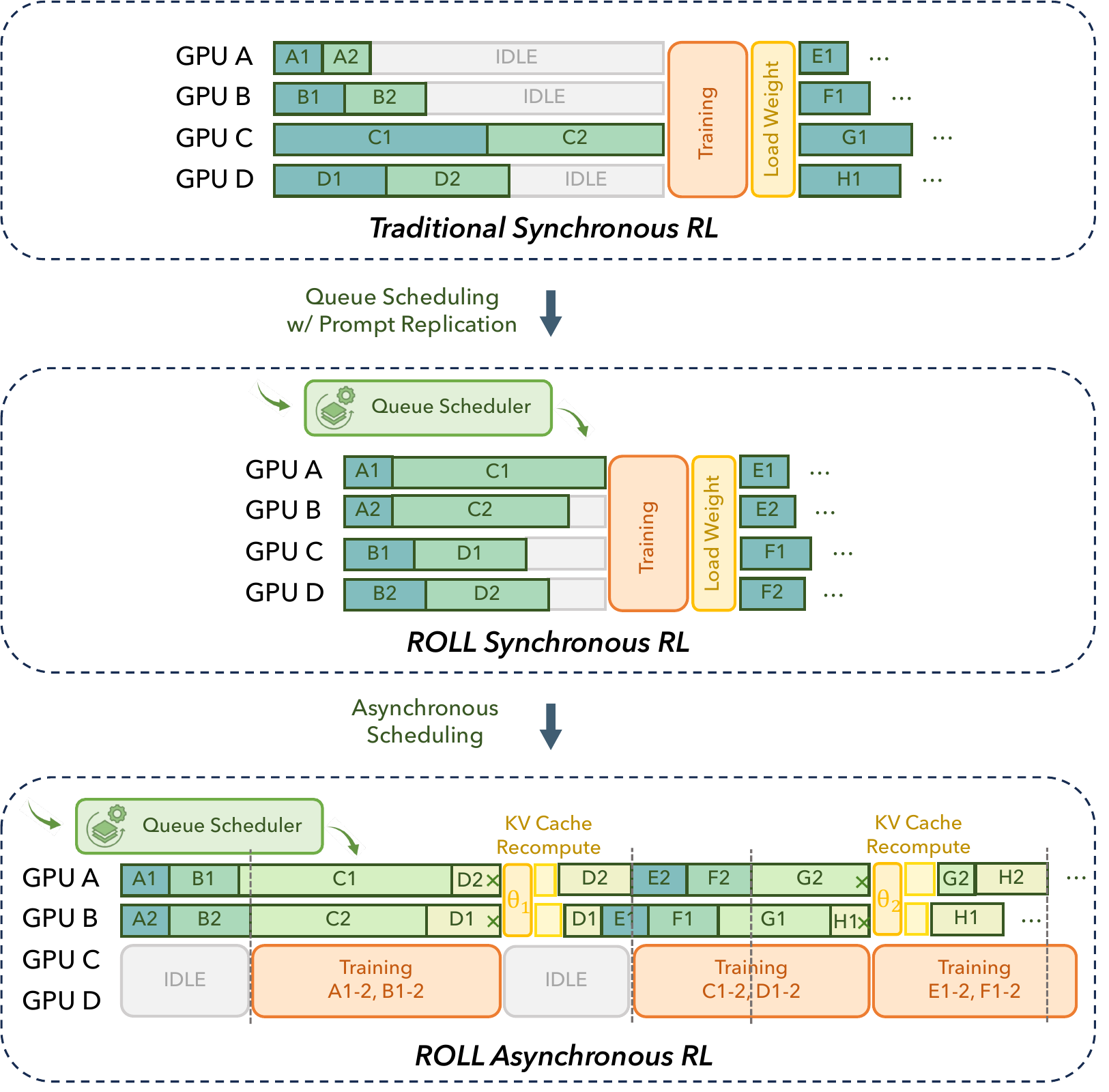}
        % \vspace{2pt}
        \caption{Overview of ROLL-Sync and Async Framework.}
        \label{fig:overview}
    \end{subfigure}%
    \hfill
    % 右侧两个子图堆叠（用 minipage 包裹）
    \begin{subfigure}[t!]{0.465\textwidth}
        \centering
        % 上图 (b)
        \includegraphics[width=\linewidth, height=0.25\textheight, keepaspectratio]{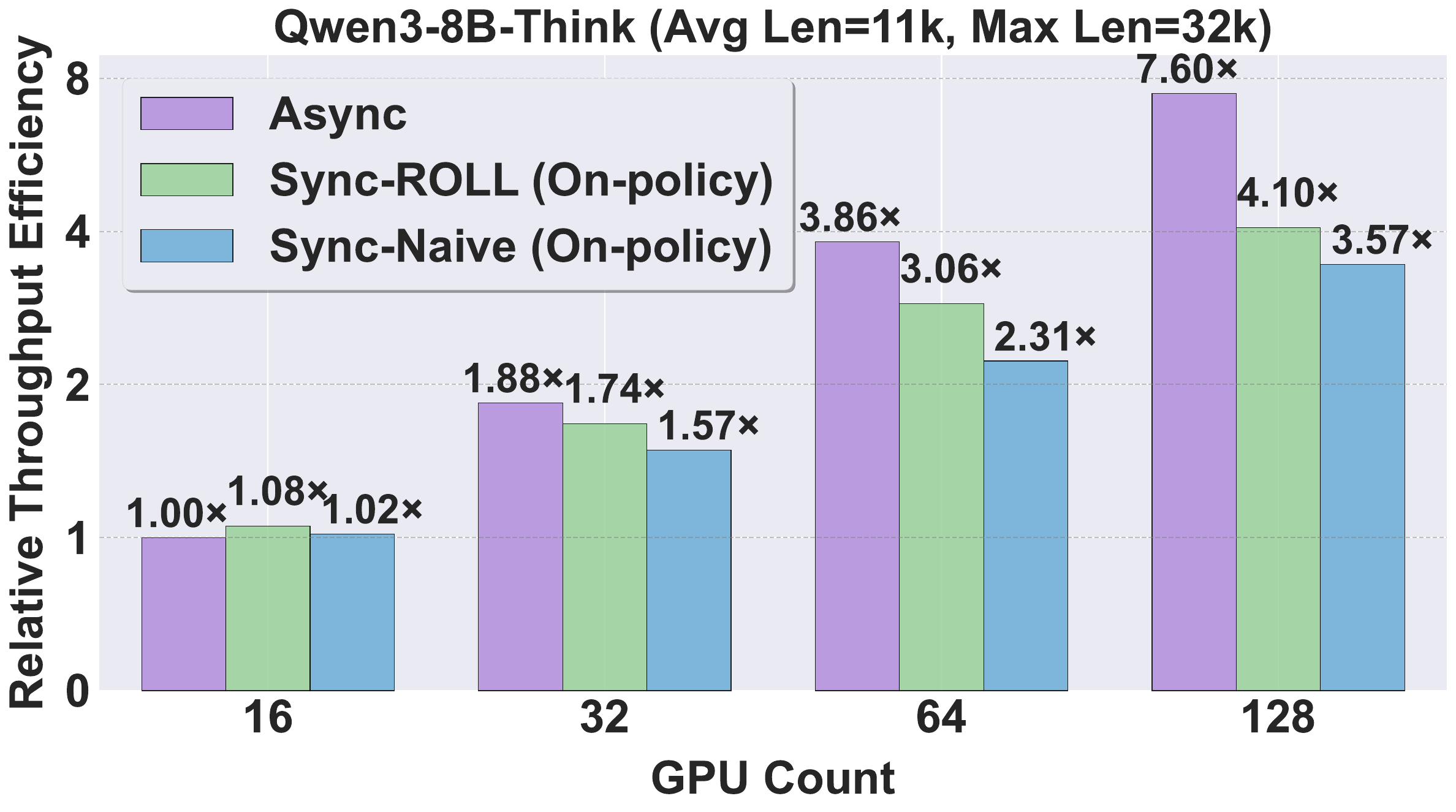}
        % \caption{Async \& Sync Scaling of GPUs}
        % \label{fig:Throughput_above}

        % \vspace{0.3em} % 控制上下子图间距
        \vspace{2pt}

        % 下图 (c)
        \includegraphics[width=\linewidth, height=0.25\textheight, keepaspectratio]{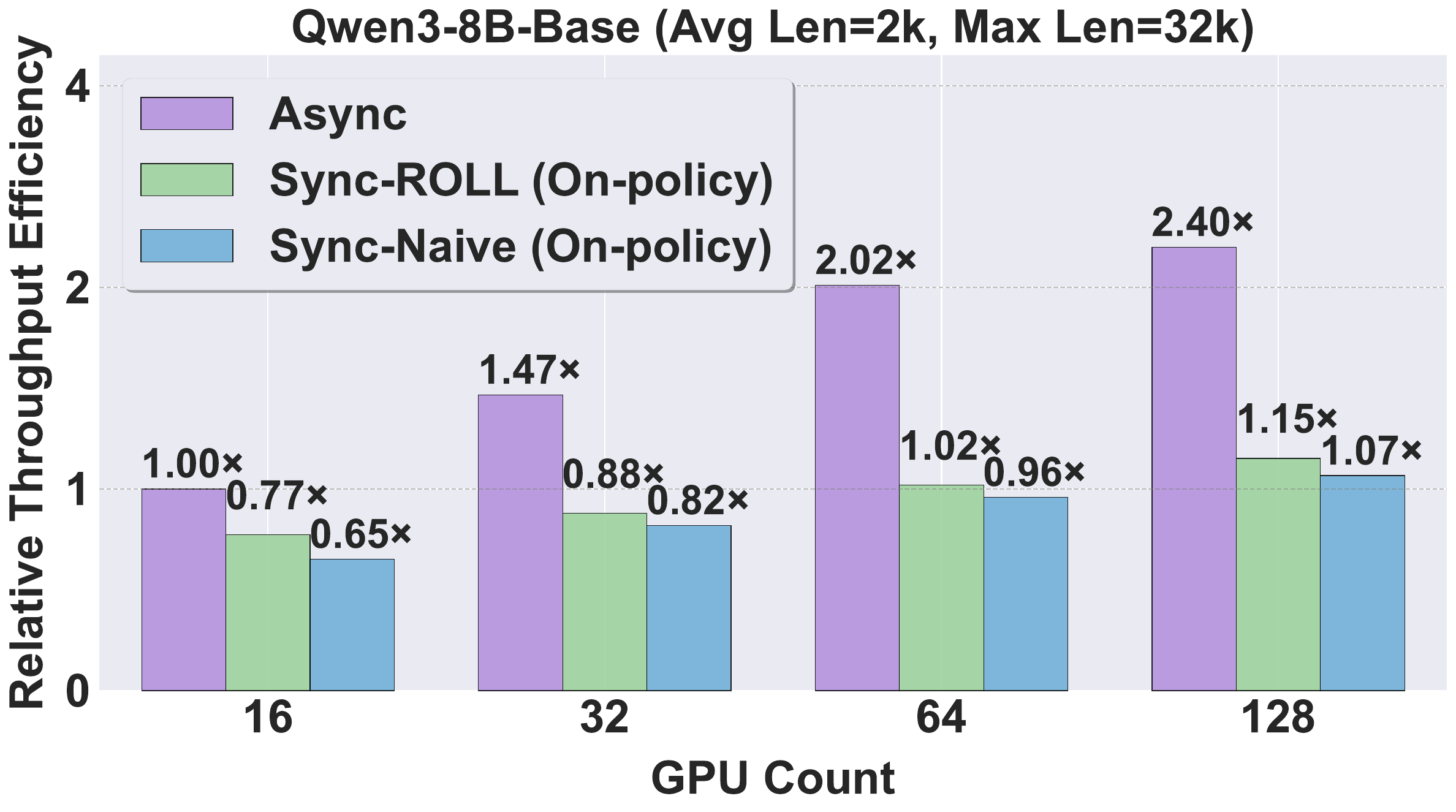} % 替换为你的第二个图
        % \vspace{-11pt}
        \caption{Throughput Efficiency Scaling with GPUs.}
        \label{fig:Throughput}
    \end{subfigure}

    % 主标题（整个 figure 的 caption）
    \caption{ \textbf{(a)} We illustrate vanilla synchronous training alongside several optimizations introduced by \SysName{}: queue scheduling (\autoref{sec:Queue_Scheduling}), prompt replication (\autoref{sec:prompt_replication}), and an asynchronous architecture (\autoref{sec:framework}). \textbf{(b)} We present how the throughput of the training architectures illustrated in~(a) scales with the number of GPUs on the \texttt{Qwen3-8B-Base} and \texttt{Think} models. In the top panel of~\autoref{fig:Throughput}, the asynchronous approach achieves higher efficiency and exhibits strong scalability with increasing GPU count, delivering $2.12\times$ throughput over synchronous structure on 128~GPUs. In the bottom of~\autoref{fig:Throughput}, all methods scale poorly at low average sequence lengths. Nevertheless, the asynchronous approach mitigates the impact of long-tail rollouts and is significantly more efficient than the synchronous approach ($1.53\times$ to $2.24\times$ faster). More detailed experiments and analyses can be found in~\autoref{sec:why_async}.} 
    \label{fig:overviewAndefficiency}
\end{figure}

\begingroup
\renewcommand{\thefootnote}{}
\footnotetext{$*$ Equal Contribution. $\dagger$ Corresponding to: Zichen Liu \texttt{<lzc410374@alibaba-inc.com>}, Shaopan Xiong \texttt{<xiongshaopan.xsp@alibaba-inc.com>} and Weixun Wang \texttt{<weixun.wwx@taobao.com>}}
\endgroup

\vfill

\newpage

\section{Introduction}
\label{sec:intro}
% rl is important and faces scability issues

Reinforcement learning (RL) has emerged as a pivotal technique for endowing large language models (LLMs) with strong reasoning capabilities in mathematics~\citep{maxwell_jia_2024_aime_dataset}, code generation~\citep{openr1_codeforces_dataset}, and tool use~\citep{pan2024trainingsoftwareengineeringagents,verl-agent} during the post-training phase. The RL post-training workflow consists of two stages, \textit{rollout} and \textit{training}, which are repeated to iteratively optimize the LLM. In the rollout stage, an actor LLM generates a batch of responses and assigns a reward signal to each response until the rollout terminates. In agentic RL tasks, the actor also interacts with the environment to produce sequences of actions and feedback that synthesize responses. In the training stage, the actor updates the model weights based on the generated responses and corresponding rewards.

Many RL post-training systems~\citep{hybridflow,realhf,openrlhf,roll,MiMo} aim to accelerate RL post-training and improve resource efficiency. Nevertheless, they often \textbf{suffer from severe resource bubbles}, particularly during the rollout stage, which accounts for over 70\% of total training time~\citep{RhymeRL,rollpacker}. Response lengths vary widely across prompts and exhibit a \emph{long-tail} distribution. The longest responses can exceed the median length by more than $20\times$~\citep{rollpacker}. A common practice is to enforce synchronization barriers between response generation, environment interaction, and reward assessment. As a result, long-tail responses lead to substantial idle time on GPUs, causing pronounced resource waste.

Moreover, existing RL post-training systems exhibit \textbf{poor resource scalability}. During the rollout stage, LLM generation performs thousands of autoregressive decoding steps to produce each complete response. Decoding is predominantly memory-bandwidth bound, so scaling out to more GPUs does not increase decoding speed. Because this decoding cost is a major contributor to end-to-end training time, adding GPUs does not substantially reduce it. The RL post-training pipeline also imposes a synchronization barrier between the rollout and training stages, i.e., the training stage begins only after rollout completes. Although additional GPUs can shorten the training compute time, they only marginally mitigate long-tail rollout overhead. Consequently, the speedup achievable through resource scaling is limited, and overall resource scalability remains poor. A seminal work AReaL~\citep{fu2025areal} presents a scalable RL post-training framework that relaxes the synchronization barrier between rollout and training. As a result, rollout proceeds continuously without blocking, and adding GPUs can scale parallelized LLM generation for more prompts. Many concurrent works~\citep{slime_github,asyncflow,deepscaler2025} also enable asynchronous training to improve the training throughput. However, asynchronous training introduces off-policy drift that can degrade model accuracy, motivating dedicated off-policy algorithms~\citep{hilton2022batch,munos2016safe,espeholt2018impala,chen2025minimax,roux2025tapered} to preserve accuracy. Thanks to combined system and algorithmic advances, asynchronous RL post-training can improve rollout throughput and resource scalability without sacrificing model performance.

In this report, we present \SysName{}, which strengthens ROLL~\citep{roll}  with \textbf{asynchronous execution}, thereby improving resource utilization and scalability for RL post-training.
% Our design is grounded in theoretical and empirical analyses of asynchronous RL post-training, which unveils that asynchrony significantly alleviates resource underutilization caused by long-tail rollouts and that throughput scales with increased GPU allocation, demonstrating a favorable resource scaling trend. 
\SysName{} satisfies two key design principles. First, \textbf{fine-grained parallelism} offers sample-level lifecycle control during the rollout stage, enabling overlap among LLM generation, environment interaction, and reward computation, thereby reducing idle time and improving GPU utilization. Leveraging this capability, we implement \texttt{prompt replication  and redundant environment rollouts}, then provide a detailed empirical analysis validating their effectiveness. Second, \textbf{rollout--train decoupling} places the rollout and training stages on separate resources and executes them in parallel. Consequently, the rollout stage does not wait for training to complete, and training can optimize the LLM using responses generated under stale policy. This decoupling is a cornerstone of asynchronous training, enabling flexible control and enhancing resource scalability. To realize these principles, \SysName{} introduces \texttt{LLMProxy}, \texttt{EnvManager}, \texttt{SampleBuffer}, and \texttt{AsyncController}. Together, these system components facilitate the implementation of the asynchronous training architecture and enable the fine-grained parallelism via queue scheduling, prompt replication, environment-level asynchronous rollout, and redundant environment rollout. 
To ensure training stability in asynchronous RL post-training, \SysName{} introduces \textbf{asynchronous ratio}, which bounds the policy version gap between the current policy and the one that initiated a sample’s generation. This per-sample freshness constraint prevents stale rollouts from degrading training performance while enabling high resource utilization.

\textbf{Theoretically,} we prove that asynchronous training is inherently more efficient than synchronous training. Asynchronous training follows a producer–consumer model, where the rollout stage remains saturated with continuous response generation and does not stall for the training stage. This effectively mitigates resource waste caused by long-tail rollouts. Practically, the resource allocation of training and rollout is coarse-grained, motivating empirical investigation for asynchronous training.

\textbf{Empirically,} we analyze across four key dimensions: resource scalability, resource utilization, asynchronous ratio, and training stability. As shown in \autoref{fig:overviewAndefficiency}, our method achieves substantial speedups over synchronous training under both \texttt{Qwen3-Base} and \texttt{Think} models~\citep{yang2025qwen3}, with gains growing consistently as GPU resources scale. Notably, \SysName{} demonstrates strong advantages in both scalability and utilization---reaching up to 2.24$\times$ higher throughput  at hundreds of GPU scale.
We further find that a small asynchronous ratio is often sufficient to realize near-maximal acceleration while preserving sample freshness. Moreover, the existing off-policy algorithms~\citep{shao2024deepseekmath,hilton2022batch,munos2016safe,chen2025minimax, roux2025tapered} can effectively compensate for potential degradation from stale samples, matching the final performance of synchronous training. 
Finally, we extend our evaluation to agentic settings, where asynchronous rollout strategies yield $2.72\times$ speedup on ALFWorld and $1.81\times$ on SWE. Together, these comprehensive experiments validate the broad effectiveness and efficiency of our approach across diverse RL and agentic workloads.

Overall, the contributions of this paper can be summarized in the following three main aspects:

\definecolor{blueviolet}{RGB}{138,43,226}
\newtcolorbox{summaryblock}{
  colback=blueviolet!5,   
  colframe=blueviolet!50!black!50!,    
  boxrule=0.5mm,       
  arc=2mm,            
  left=0pt,           
  right=8pt,           
  top=8pt,            
  bottom=8pt}

\begin{summaryblock}
\begin{enumerate}[leftmargin=1.5em]
    \item \textbf{ROLL Flash Structure}:  A system design enabling fine-grained parallelism and rollout–train decoupling, which not only supports async training but also boosts async generation efficiency.
    
    \item \textbf{Async--Sync Analysis}: Theoretical and empirical characterization of when asynchronous training excels, revealing \texttt{ROLL-Async}'s strong resource scalability and high utilization.

    \item \textbf{RLVR \& Agentic Acceleration}: In comprehensive experiments and ablation study, \texttt{ROLL-Flash} delivers substantial speedups, up to \textbf{$2.24\times$} in RLVR tasks and \textbf{$2.72\times$} in agentic tasks. 
\end{enumerate}
\end{summaryblock}

\begin{figure}[h]
    \centering

    % ============= 第一行：单个宽图（无 subcaption） =============
    \includegraphics[width=0.8\textwidth]{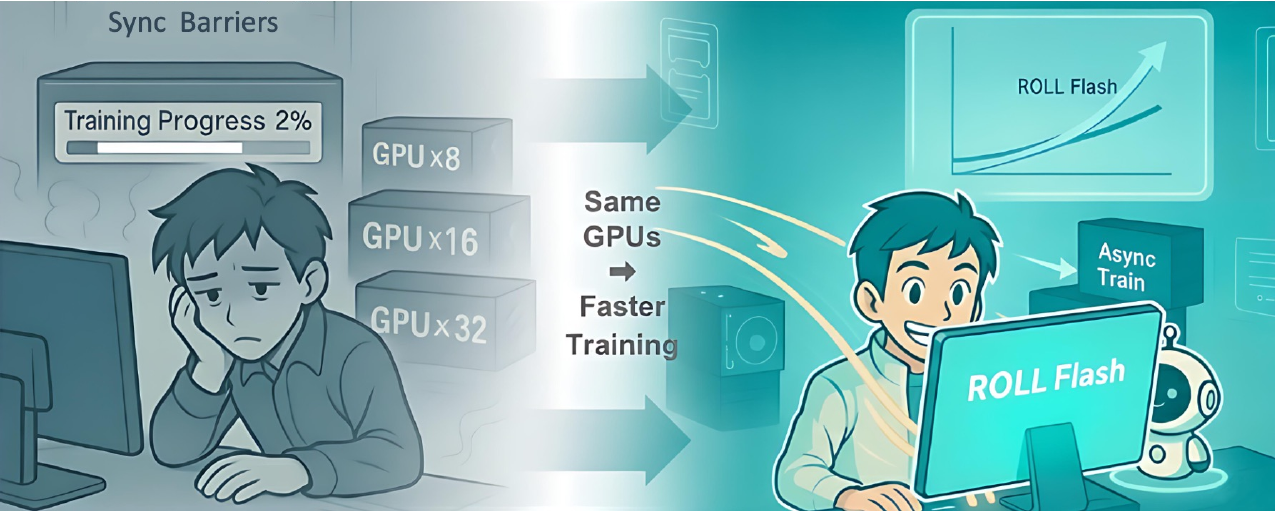}
    
    % \vspace{-1mm} % 增加一点垂直间距，避免拥挤

    % ============= 第二行：左右两个子图组 =============
    % \includegraphics[width=0.9\textwidth]{figures/RLVR/Async_Entropy_ResponseLength.pdf}
    % \vspace{-3mm}
    \caption{\textbf{An illustration of Training Acceleration with \SysName{}.}}
    \label{fig:gen_images}
\end{figure}

% \begin{figure}[h]
%     \centering

%     % ============= 第一行：单个宽图（无 subcaption） =============
%     \includegraphics[width=\textwidth]{figures/gen_images.png}
    
%     % \vspace{-1mm} % 增加一点垂直间距，避免拥挤

%     % ============= 第二行：左右两个子图组 =============
%     % \includegraphics[width=0.9\textwidth]{figures/RLVR/Async_Entropy_ResponseLength.pdf}
%     % \vspace{-3mm}
%     \caption{\textbf{1}} 
%     \label{fig:gen_images}
% \end{figure}

% long-tail rollouts is the key of 
% We propose a holistic asynchronous solution to improve the scability of RL post-training. 

% Asynchronous is a key effective: 

% async execution LLM generation with environment execition 

% async rollouts with rewards 

% async execution rollouts with the training 

% 

% \begin{insightblock}
% \begin{enumerate}[leftmargin=1.5em]
%     \item xxx. (\S\ref{})
% \end{enumerate}
% \end{insightblock}

% \begin{tcolorbox}[colback=cyan!5!white, colframe=cyan!45!blue!60, title=\textbf{Feature 1}]

% \end{tcolorbox}
\section{Background and Preliminaries}
\label{sec:background}

% 2.1, definition 
% 2.2, the challenge: e.g., async 
% 2.3. async algorithm
\subsection{Synchronous RL Post-Training}
The RL post-training comprises three stages: rollout, reward, and training. We use an agentic task to illustrate the workflow. During rollout, the agent LLM interacts with environments over multiple turns, producing tuples of states and actions that form a trajectory. Then, a reward worker assigns each trajectory a score. Last, the LLM updates its weights using these trajectories and rewards during training.

The synchronous training requires strict synchronization of the model weights in each training step, thus creating barriers between the rollout stage and the training stage, leading to substantial resource bubbles and underutilization. Many algorithms are employed to maximize the learning efficiency of RL post-training. Two representative examples are PPO and GRPO.

\paragraph{Proximal Policy Optimization (PPO).}
PPO \citep{schulman2017proximal} is a widely used policy gradient algorithm based on the actor-critic framework. It enhances stability by optimizing a clipped surrogate objective, which restricts how much the updated policy $\pi_\theta$ can deviate from the old policy $\pi_{\theta_{\mathrm{old}}}$ at each update step. The objective is defined as:

\begin{equation}
\begin{aligned}
\mathcal{J}_{\mathrm{PPO}}(\theta) =\ 
&\mathbb{E}_{\left[ 
  q \sim P(Q),\ 
  o \sim \pi_{\theta_{\mathrm{old}}}(O|q)
\right]} \\
&
  \frac{1}{|o|} \sum_{t=1}^{|o|}
  \min\Bigg(
    \frac{\pi_\theta(o_t|q, o_{<t})}{\pi_{\theta_{\mathrm{old}}}(o_t|q, o_{<t})} A_t,\, 
    \mathrm{clip}\left(
      \frac{\pi_\theta(o_t|q, o_{<t})}{\pi_{\theta_{\mathrm{old}}}(o_t|q, o_{<t})},\, 1{-}\epsilon,\, 1{+}\epsilon
    \right) A_t
  \Bigg),
\end{aligned}
\end{equation}

where $\pi_\theta$ and $\pi_{\theta_{\text{old}}}$ represent the current and previous policies, respectively. Here $q$ denotes a sampled question, $o$ is the generated sequence, with $o_t$ representing the $t$-th token in $o$ and advantage $A_t$, typically computed via Generalized Advantage Estimation (GAE)~\citep{schulman2015high}; $\epsilon$ is the hyperparameter controlling the clipping range. The combination of min and clip ensures that, whether the advantage is positive or negative, the policy update remains controlled, thereby maximizing positive rewards while suppressing overadjustments that could lead to negative outcomes, ultimately maintaining the stability and effectiveness of the learning process.

\paragraph{Group Relative Policy Optimization (GRPO).} Despite PPO's robustness across tasks, its reliance on a critic reveals limitations in language generation: advantage estimation becomes unstable under sparse rewards for long sequences, and unreliable value estimation may exacerbate policy convergence in suboptimal landscapes while incurring  non-negligible extra computational overhead. To address this, Group Relative Policy Optimization (GRPO)~\citep{shao2024deepseekmath} proposes a critic-free alternative that constructs advantage signals by sampling multiple responses per prompt and normalizing their rewards. Specifically, given a prompt $q$ and $G$ output sequences with rewards $\{r_i\}_{i=1}^G$, GRPO defines the normalized advantage for the $t$-th token of the $i$-th sequence as:
\begin{equation}
\hat{A}_{i,t} = \frac{r_i - \mathrm{mean}(\{r_i\}_{i=1}^G)}{\mathrm{std}(\{r_i\}_{i=1}^G)}.
\label{eq:grpo_norm_final}
\end{equation}

GRPO standardizes rewards via group statistics (mean and std), enabling meaningful learning signals through relative ranking—even under sparse or similar rewards. Theoretically, it acts as reward shaping that emphasizes intra-group differences to preserve gradient discriminability~\citep{hu2020learning}. GRPO adds KL divergence explicitly as a regularization term in the loss. The objective function is
\begin{equation}
\begin{aligned}
\mathcal{J}_{\text{GRPO}}(\theta) = &\ \mathbb{E}_{q \sim P(Q),\ \{o_i\}_{i=1}^G \sim \pi_{\theta_{\text{old}}}(O|q)} \\&  \frac{1}{G} \sum_{i=1}^G \frac{1}{|o_i|} \sum_{t=1}^{|o_i|} \bigg\{ \min \bigg( r_{i,t}(\theta)\, \hat{A}_{i,t},\ \text{clip}\big(r_{i,t}(\theta),\ 1-\epsilon,\ 1+\epsilon\big)\, \hat{A}_{i,t} \bigg) - \beta D_{\mathrm{KL}}[\pi_\theta \parallel \pi_{\text{ref}}] \bigg\} ,
\end{aligned}
\label{eq:grpo_final}
\end{equation}
where $r_{i,t}(\theta) = \frac{\pi_\theta(o_{i,t}|q,o_{i,<t})}{\pi_{\theta_{\text{old}}}(o_{i,t}|q,o_{i,<t})}$ and $\beta$ is the regularization weight, and $D_{\mathrm{KL}}$ denotes the KL divergence between current policy and the reference policy.

\subsection{Asynchronous LLM Post-Training}

As observed in prior RL post-training systems~\citep{rollpacker,RhymeRL}, the rollout stage typically accounts for over 70\% of the total training time, and long-tail rollouts incur substantial resource idleness and underutilization. Moreover, scaling out GPUs does not mitigate these long-tail rollouts and yields only marginal computational gains. The asynchronous training emerges as a promising technique to alleviate this issue and improve the resource utilization and scalability~\citep {mnih2016asynchronous,wu2025llamarl,fu2025areal}. Asynchronous training decouples the rollout and training stages and runs them in parallel on separate resources, eliminating the strict synchronization barrier. In other words, the training stage can consume responses generated using stale model weights from the rollout stage, while the rollout stage concurrently produces new responses without waiting for model updates.

While asynchronous training can improve computational efficiency, it introduces policy staleness that can degrade model accuracy~\citep{fu2025areal}, underscoring the need for dedicated off-policy algorithmic support. Specifically, responses are sampled from the old distribution $\pi_{\mathrm{old}}$, which typically differs from the current policy $\pi_\theta$. When we apply synchronous training algorithms (e.g., PPO) in this asynchronous setting, the divergence between the old and current policies can induce policy collapse~\citep{chen2023sufficiency} and bias the gradients, leading to training instability and severe performance degradation. To restore accuracy, we adopt off-policy training algorithms to stabilize the training dynamics. The mainstream off-policy RL post-training algorithms typically fall into two categories: (1) \textbf{Gradient truncation}, which truncates gradients for tokens whose importance-sampling (IS) ratios lie outside a trust region; e.g., Decoupled PPO~\citep{hilton2022batch}. (2) \textbf{Importance-sampling optimization}, which retains gradients for all samples but clips the importance sampling weights to stabilize training; e.g., Truncated IS~\citep{munos2016safe,espeholt2018impala}, CISPO~\citep{chen2025minimax} and TOPR~\citep{roux2025tapered}.

\noindent\textbf{Notation.} In the following formulations, $R(\tau)$ denotes the learning signal associated with trajectory $\tau$, which can also be advantage estimate $A(\tau)$ in practice. We use $\mathbf{sg}(\cdot)$ to denote the stop-gradient operator (gradients are not backpropagated through this term) and $\mathbf{1}_{\{\cdot\}}$ for the indicator function. The shorthand $(x)^{a}_{b}$ denotes $\mathrm{clip}(x, b, a)$, i.e., $x$ is constrained to lie between a lower bound $b$ and an upper bound $a$.

\definecolor{blueviolet}{RGB}{138,43,226}
\newtcolorbox{algoblock}{
  % colback=cyan!5!white,
  % colframe=cyan!45!blue!60,
  boxsep=0pt,
  top=2pt,
  % attach title to upper=\par,
  title=\textbf{Loss Objective for Off-policy Algorithms}, }
\begin{algoblock}
\begin{align*}
\textbf{PPO (Standard)}:\quad
& \mathcal{J}^{\mathrm{PPO}}(\pi_\theta)
= \mathbb{E}_{\tau \sim \pi_{\mathrm{old}}}
\left[
\min \left(
R(\tau)\frac{\pi_\theta(\tau)}{\pi_{\mathrm{old}}(\tau)},
R(\tau)\left(\frac{\pi_\theta(\tau)}{\pi_{\mathrm{old}}(\tau)}\right)^{1+\epsilon}_{1-\epsilon}
\right)
\right]
% \label{eq:ppo}
\\[6pt]
\textbf{Decoupled PPO}:\quad
& \mathcal{J}^{\mathrm{DPPO}}(\pi_\theta)
= \mathbb{E}_{\tau \sim \pi_{\mathrm{old}}}
\left[
\min \left(
R(\tau)\frac{\pi_\theta(\tau)}{\pi_{\mathrm{old}}(\tau)},
R(\tau)\textcolor{MediumPurple}{\frac{\pi_{\mathrm{prox}}(\tau)}{\pi_{\mathrm{old}}(\tau)}}\left(\textcolor{MediumPurple}{\frac{\pi_\theta(\tau)}{\pi_{\mathrm{prox}}(\tau)}}\right)^{1+\epsilon}_{1-\epsilon}
\right)
\right]
% \label{eq:decoupled-ppo}
\\[6pt]
\textbf{Truncated IS}:\quad
& \mathcal{J}^{\mathrm{TIS}}(\pi_\theta)
= \mathbb{E}_{\tau \sim \pi_{\mathrm{old}}}
\left[\textcolor{purple}{\mathbf{sg}}
\left(\frac{\pi_\theta(\tau)}{\pi_{\mathrm{old}}(\tau)}\right)^{\textcolor{blue}{c}}_{0}
R(\tau)\, \textcolor{purple}{\log \pi_\theta(\tau)}
\right]
% \label{eq:tis}
\\[6pt]
\textbf{CISPO}:\quad
& \mathcal{J}^{\mathrm{CISPO}}(\pi_\theta)
= \mathbb{E}_{\tau \sim \pi_{\mathrm{old}}}
\left[ \textcolor{purple}{\mathbf{sg}}
\left(\frac{\pi_\theta(\tau)}{\pi_{\mathrm{old}}(\tau)}\right)^{1+\epsilon^{\mathrm{IS}}_{\mathrm{high}}}_{1-\epsilon^{\mathrm{IS}}_{\mathrm{low}}}
R(\tau)\, \textcolor{purple}{\log \pi_\theta(\tau)}
\right]
% \label{eq:cispo}
\\[6pt]
\textbf{TOPR}:\quad
& \mathcal{J}^{\mathrm{TOPR}}(\pi_\theta)
= \mathbb{E}_{\tau \sim \pi_{\mathrm{old}}} \left[
\Big(
\textcolor{violet}{\mathbf{1}_{\{\tau \in T^+\}}}
+ \textcolor{violet}{\mathbf{1}_{\{\tau \in T^-\}}} \ \textcolor{purple}{\mathbf{sg}}
\left(\tfrac{\pi_\theta(\tau)} {\pi_{\mathrm{old}}(\tau)}\right)^{\textcolor{blue}{c}}_{0}
\Big) R(\tau)\, \textcolor{purple}{\log \pi_\theta(\tau)}
\right]
% \label{eq:topr}
\end{align*}
\end{algoblock}

Decoupled PPO introduces a proximal policy $\pi_{\mathrm{prox}}$ to better regulate policy updates. TIS and TOPR both employ a truncation threshold $c$ to cap the importance sampling ratio from above, mitigating variance and instability.  
In contrast, PPO and CISPO constrain the ratio within a symmetric or asymmetric interval around 1, controlled by $\epsilon^{\mathrm{IS}}_{\mathrm{low}}$ and $\epsilon^{\mathrm{IS}}_{\mathrm{high}}$. Notably, TOPR partitions trajectories into two sets: $T^+$ (high-return/correct) and $T^-$ (low-return/incorrect), applying truncation only to $T^-$ to preserve learning signals from good trajectories while suppressing noise from poor ones. \SysName{} has integrated the above off-policy algorithms to facilitate performance of asynchronous training.

% \clearpage
\section{Performance-Preserving Asynchronous Acceleration}
\label{sec:why_async}

\subsection{Theoretical Analysis}  

In the \SysName{} architecture, we adopt two designs: (1) Queue Scheduling with Prompt Replication (\autoref{sec:RLVR}), where responses are scheduled individually and immediately on any idle worker; (2) in Async, the same total number of GPUs is partitioned between training and inference. The \emph{asynchrony ratio} $\alpha$ (\autoref{sec:async_ratio}) denotes how many model updates the rollout policy is allowed to lag behind the current training model, and it directly determines the size of the generation data pool.

\begin{proposition}{Generation Time Bound}{gentime}
Let there be $K$ workers executing in a Queue Scheduling manner (a new task is assigned immediately once a worker finishes). Suppose $Q$ samples need to be generated, where the generation time of each sample lies in $[0, L_{\text{gen}}]$ with mean $\mu_{\text{gen}}$. Then the total completion time satisfies:
\begin{equation}
T_{\text{completion}} \leq \frac{Q}{K} \mu_{\text{gen}} + L_{\text{gen}}.
\end{equation}
Consequently, the average completion time per sample is bounded by:
\begin{equation}
\frac{T_{\text{completion}}}{Q} \leq \frac{\mu_{\text{gen}}}{K} + \frac{L_{\text{gen}}}{Q}. \label{eq:completion_time}
\end{equation}
% Specifically:
\begin{itemize}
    \item In the \textbf{Sync} setting ($Q = N$), the average per-sample completion time satisfies:
    \begin{equation}
    \overline{T}_{\text{sync}} \leq \frac{\mu_{\text{gen}}}{K} + \frac{L_{\text{gen}}}{N}.
    \end{equation}
    
    \item In the \textbf{Async} setting ($Q = (\alpha + 1) N$, with $\alpha
    $ denoting the \emph{asynchrony ratio} (details in \S\ref{sec:async_ratio})):
    \begin{equation}
    \overline{T}_{\text{async}} \leq \frac{\mu_{\text{gen}}}{K} + \frac{L_{\text{gen}}}{(\alpha + 1) N}.
    \end{equation}
\end{itemize}

As $\alpha \to \infty$, the per-sample completion time converges to $\mu_{\text{gen}} / K$.  
When $K = N$, the maximum theoretical speedup achievable by Async over Sync setting is at most $(L_{\text{gen}} + \mu_{\text{gen}}) / \mu_{\text{gen}}$.
\end{proposition}

\begin{proposition}{End-to-End Efficiency with Resource Partitioning}{H}
Consider a system with $K$ workers. The resource allocation strategy are as follows: 
\begin{itemize}
\item \textbf{Sync}: All $K$ workers generate $N$ samples and train sequentially afterward.
    \item \textbf{Async}: Workers are partitioned into two disjoint pools controlled by a parameter $\beta \in (0,1)$:
    \begin{itemize}
        \item $(1 - \beta)K$ workers are allocated to continuous sample generation,
        \item $\beta K$ workers are used for parallel training.
    \end{itemize}
\end{itemize}
Let $\mu_{\text{gen}}$ denote the average sample generation time (maximum $L_{\text{gen}}$), $\mu_{\text{train}}$ denote the average training time per sample. Define $E$ as the number of each generated sample reused during training.
\begin{enumerate}
    \item The end-to-end completion time for \textbf{Sync} (sequential pipeline) is:
    \begin{equation}
    T_{\text{sync}} \leq \frac{N}{K} \mu_{\text{gen}} + L_{\text{gen}} + E \frac{N}{K} \mu_{\text{train}} = \frac{N}{K}(\mu_{\text{gen}} + E \mu_{\text{train}}) + L_{\text{gen}}.
    \label{eq:sync_end2end}
    \end{equation}

    \item The end-to-end completion time for \textbf{Async} (parallel, resource-isolated pipeline) is:
    \begin{equation}
    T_{\text{async}} \leq \max\left( 
        \frac{N}{(1-\beta)K} \mu_{\text{gen}} + \frac{L_{\text{gen}}}{(\alpha + 1)(1-\beta)},\ 
        \frac{E N}{\beta K} \mu_{\text{train}} 
    \right).
    \label{eq:async_end2end}
    \end{equation}
\end{enumerate}

The optimal worker allocation ratio $\beta^*$ that minimizes the upper bound on $T_{\text{async}}$ is 
% obtained by equating the two terms inside the $\max$:
\begin{equation}
    \beta^* = \frac{E N \mu_{\text{train}}}{N \mu_{\text{gen}} + \dfrac{K L_{\text{gen}}}{\alpha + 1} + E N \mu_{\text{train}}}.
\end{equation}

At this optimal $\beta^*$, the two components are balanced, and the resulting upper bound becomes:
\begin{equation}
T_{\text{async}} \leq \frac{N}{K} (\mu_{\text{gen}} + E \mu_{\text{train}}) + \frac{L_{\text{gen}}}{\alpha + 1}.
\end{equation}
\textbf{The Async setting yields a tighter theoretical bound and strictly improves upon it when} $\alpha > 0$. 
As $\alpha \to \infty$, the maximum theoretical speedup of Async over Sync converges to $1 + \frac{KL_{\text{gen}}}{N(\mu_{\text{gen}} + E\mu_{\text{train}})}$.
\end{proposition}

\subsection{Experimental Validation} 

We further provide a detailed empirical analysis of asynchronous training, revealing four key takeaways: \textbf{resource scalability, resource utilization, asynchronous ratio, and training stability}.

\textbf{Experimental Setup.} Unless otherwise specified, all experiments in this section are conducted using the \texttt{Qwen3-8B-Base} or \texttt{Think} models~\citep{yang2025qwen3}, with a sequence length of 32k, 256 rollouts, and a group size of 32 prompts per minibatch on the DAPO-Math-18K~\citep{yu2025dapo} dataset (other details in \autoref{app:training_details}).  In the following experiments, we conduct a detailed ablation study by varying several key parameters, including: (1) the choice of base model—either \texttt{Qwen3-8B-Base} (average length 2k) or \texttt{Qwen3-8B-Think} (average length 11k); (2) the number of GPUs, ranging from 16 to 128; (3) the rollout batch size, varied from 32 to 512; and (4) different GPU allocation ratios between training and inference.

In \autoref{fig:Throughput}, the evaluated paradigms include:
(1) \textbf{Async}: ROLL’s asynchronous architecture with an Async Ratio of 2;
(2) \textbf{Sync-ROLL (On-policy)}: A synchronous architecture enhanced with ROLL-specific optimizations, including Queue Scheduling and Prompt Replication;
(3) \textbf{Sync-Naive (On-policy)}: A standard synchronous reinforcement learning setup. The default training-to-inference GPU ratio is 1:1.

% (4) \textbf{Sync-Naive (RolloutBS \(\times\) 4)}: A variant that generates four minibatches per generation step to mitigate the long-tail latency effect, at the cost of introducing inherent off-policy behavior. 

\begin{tcolorbox}[
  colback=cyan!5!white,
  colframe=cyan!45!blue!60,
  title=\textbf{Takeaway 1: Async Architecture Achieves Superior Throughput Scalability.}
]
Increasing GPU resources causes Sync to suffer more from the impact of long-tail samples, whereas Async exhibits better scaling behavior and achieves higher resource utilization.
\end{tcolorbox}

\autoref{fig:Throughput} illustrates the throughput efficiency of different training paradigms as the number of GPUs scales. Under the \texttt{Qwen3-8B-Think} model, the asynchronous (\textbf{Async}) approach achieves near-linear throughput scaling with GPU resources—reaching an \textbf{impressive $7.6\times$ speedup with $8\times$ GPUs}, which is \textbf{$2.13\times$ higher} than that of the traditional synchronous (\textbf{Sync-Naive}) baseline. Under the \texttt{Qwen3-8B-Base} model where the average generation length is significantly shorter, the system is no longer compute-bound. Consequently, resource utilization drops substantially across all architectures. While throughput under synchronous methods (\textbf{Sync-Naive} and \textbf{Sync-ROLL}) plateaus, the \textbf{Async} architecture continues to scale effectively, \textbf{achieving $2.24\times$ higher throughput} than \textbf{Sync-Naive} at 128 GPUs.

This behavior arises because increasing the number of GPUs reduces the per-GPU workload (i.e., rollouts per device), thereby amplifying the impact of long-tail effects. This issue is further exacerbated in the \texttt{Base} setting, where response lengths exhibit high variance. Traditional synchronous training suffers severe efficiency degradation under such conditions, while our \textbf{Sync-ROLL} variant partially alleviates the problem through ROLL-specific optimizations (e.g., queue scheduling and prompt replication). 

In contrast, the \textbf{Async} architecture fundamentally eliminates the straggler bottleneck by decoupling generation and training. These results lead to a clear conclusion: \textbf{in resource-rich regimes or scenarios with pronounced long-tail generation latency, the asynchronous architecture enables significantly more efficient resource utilization and should be the preferred choice.}

\begin{tcolorbox}[colback=cyan!5!white, colframe=cyan!45!blue!60, title=\textbf{Takeaway 2: Async Accelerates Training in Almost All Cases.}]
Async effectively mitigates training stalls caused by long-tail generation latencies, delivering substantial speedups when the allocation of training and inference resources is well balanced.
\end{tcolorbox}

\begin{figure}[htbp]
    \centering
    \begin{subfigure}[t]{0.49\textwidth}  % ← [t] 顶部对齐
    \centering
    \includegraphics[width=\linewidth, height=6cm, keepaspectratio]{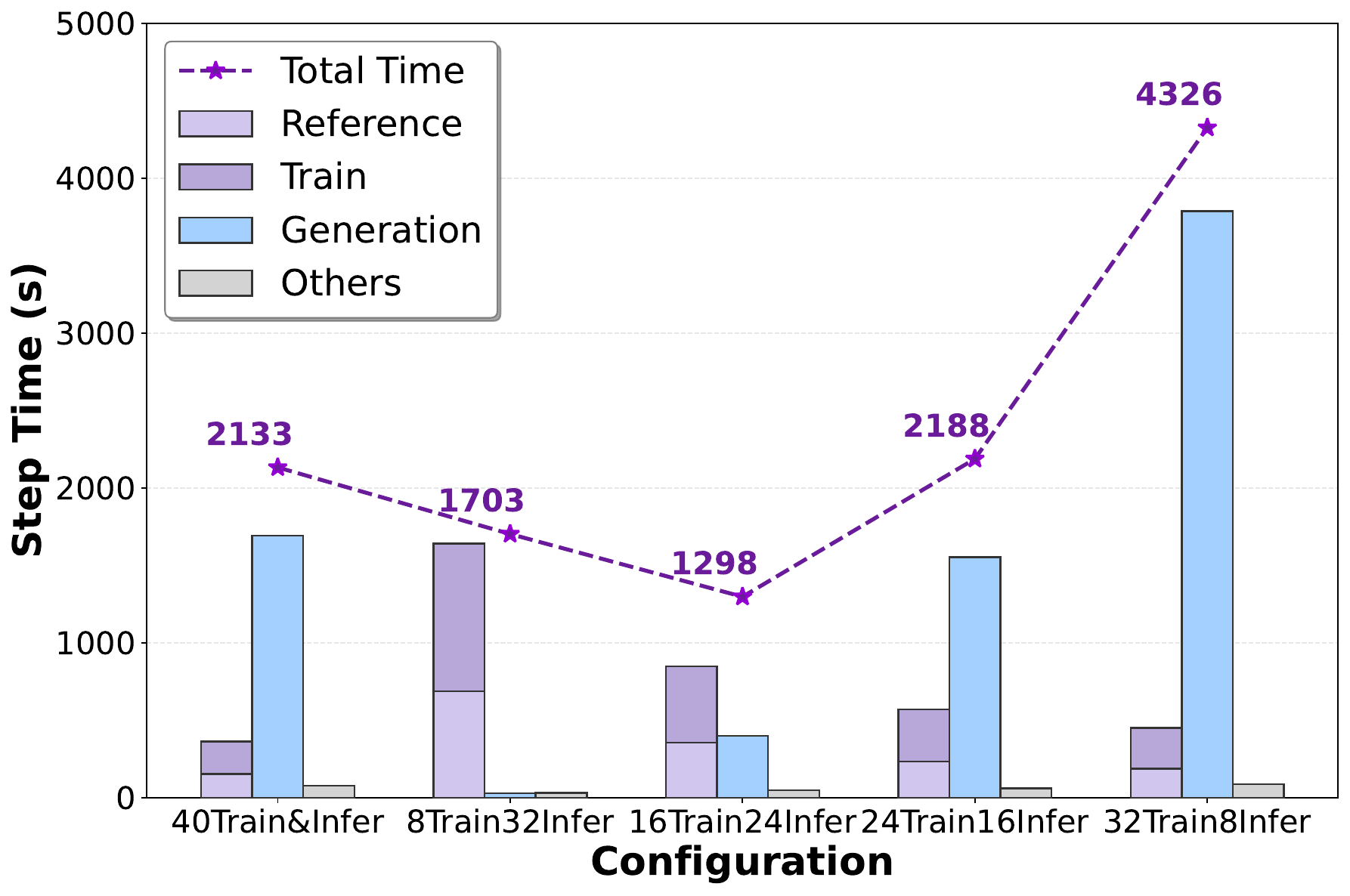}
    \caption{Time across Training-Inference Resource Allocation.}
    \label{fig:Async_Async_Efficient_figa}
    \end{subfigure}
    \hfill
    \begin{subfigure}[t]{0.49\textwidth}  % ← [t] 顶部对齐
        \centering
        \includegraphics[width=\linewidth, height=6cm, keepaspectratio]{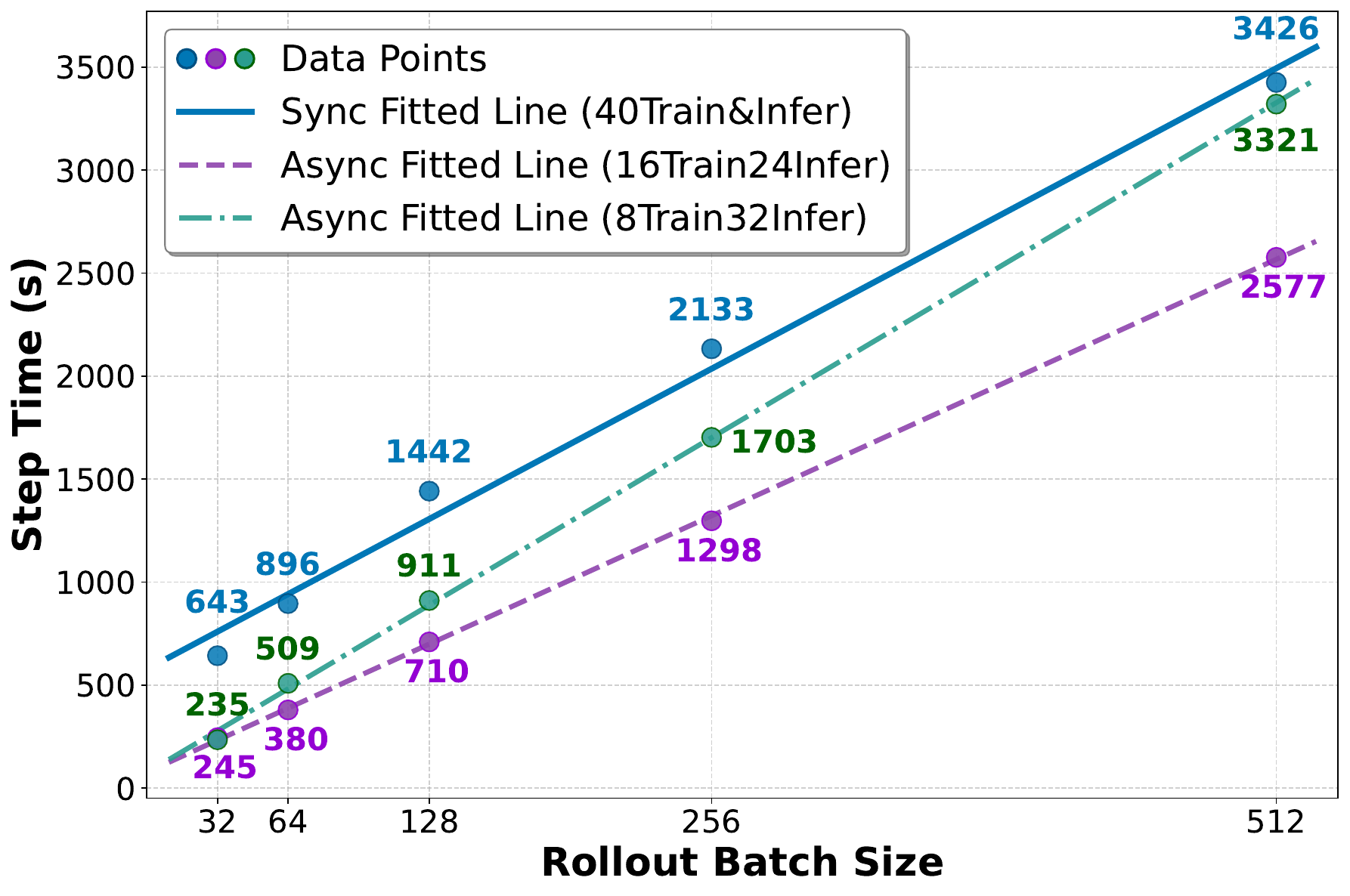}
        \caption{Async \& Sync Scaling of Step Time vs. Rollout Size.}
        \label{fig:Async_Async_Efficient_figb}
    \end{subfigure}
    \caption{\textbf{Efficiency Comparison using Async and Sync under different rollout batch size and training-inference resource ratios.} (a) Given a fixed GPU resource budget, optimal efficiency can be achieved by tuning the allocation ratio between training and inference. (b) shows the efficiency scaling curves of Async and the ROLL-Sync. Async exhibits a clear advantage in almost all cases. }
    \label{fig:Async_Async_Efficient}
\end{figure}

\textbf{The core acceleration benefit of Async stems from eliminating resource waste and idle waiting caused by long-tail generation latencies.} In an idealized scenario, the consumption rate (training) should roughly match the production rate (generation), with the Async Ratio serving to absorb tail latency and prevent training stalls. A critical design decision, therefore, is to optimally allocate resources between training and inference to maximize overall efficiency. 

\autoref{fig:Async_Async_Efficient_figa} illustrates efficiency gains under varying Train-Inference resource allocations. A well-tuned Async configuration (16 GPUs for training and 24 GPUs for inference) achieves nearly $2\times$ speedup over baseline\footnote{Notably, to support various off-policy methods and evaluation metrics, our training phase includes not only parameter updates but also inference passes over both the initial and proximal reference models. }. 
In contrast, ROLL-Sync spends substantial time waiting for sample generation. Under the 32Infer setting, although training never needs to wait for data generation, computational resources are excessively underutilized during generation. In contrast, the 24Infer configuration achieves the best overall performance. \textbf{Interestingly, a modest amount of waiting for freshly generated samples not only avoids waste but also helps stabilize training by enabling the use of more up-to-date data.}

\autoref{fig:Async_Async_Efficient_figb} shows the per-step training time of Sync and Async as a function of rollout batch size. For a fixed number of samples, training time scales approximately linearly with sample count, with fixed constant overheads such as model loading and offloading. As predicted by~\autoref{eq:completion_time}, generation time is governed by the interplay between average and tail-case latencies, and the observed step times indeed exhibit near-linear scaling. The slope of each curve reflects the marginal cost of processing additional samples. The rollout sizes in~\autoref{fig:Async_Async_Efficient_figb} already correspond to realistic on-policy or off-policy training regimes. As further confirmed by~\autoref{fig:overviewAndefficiency}, Async scales more favorably than Sync with increasing GPU count. \textbf{Therefore, Async can accelerate training in nearly all practical scenarios.}

\begin{tcolorbox}[colback=cyan!5!white, colframe=cyan!45!blue!60, title=\textbf{Takeaway 3: Async Ratio Can Be Small Enough.}]
In typical configurations, setting the Asynchronous Ratio to 2 achieves the highest throughput, effectively balancing learning efficiency and the degree of off-policy learning.
\end{tcolorbox}

\begin{wraptable}{r}{0.55\textwidth} % r = right, 宽度 0.5\textwidth
    \centering
    \vspace{-10pt}
    \caption{\textbf{Async Ratio Required in various Configuration.}}
    \label{tab:async_ratio}
    \vspace{-5pt}
    \resizebox{0.9\linewidth}{!}{% ← 缩放到 wraptable 的宽度（推荐用 \linewidth）
    \begin{tabular}{l|cccc}
        \midrule[1pt]
        \rowcolor{lightgray}
        \textbf{Model Size} & 0.6B & 1.7B & 4B & 8B \\
        Async Ratio & 2 & 2 & 2 & 2 \\
        \midrule
        \rowcolor{lightgray}
        \textbf{Length} & 4K & 8K & 16K & 32K \\
        Async Ratio & 1 & 1 & 1 & 2 \\
        \midrule
        \rowcolor{lightgray}
        \textbf{Rollout Size} & 32 & 64 & 128 & 256 \\
        Async Ratio & 4 & 2 & 2 & 2 \\
        \bottomrule[1pt]
    \end{tabular}
    }

    \vspace{-15pt}
\end{wraptable}

In Async architecture, the Async Ratio is a critical hyperparameter. If set too low, sample generation may lag behind training, causing long-tail samples as a bottleneck that limits overall throughput. Conversely, if set too high, training samples become excessively stale, degrading both training stability and effectiveness due to outdated policy sampling.

We aim to \textbf{identify the optimal Async Ratio that maximizes throughput.} To this end, we evaluate under a standard configuration: \texttt{Qwen3-8B-Think}, sequence length of 32K, and rollout batch size of 256. \textbf{The optimal Async Ratio depends on generation throughput.} In a 32Train8Infer setup, an Async Ratio of \textbf{1} yields a 25\% throughput improvement over the fully synchronous baseline—the highest achievable in this setting. In contrast, under a 24Train16Infer configuration, an Async Ratio of 1 provides a 28\% speedup, while a ratio of \textbf{2} unlocks the full benefit, delivering a 64\% throughput gain. Further increasing the Async Ratio beyond this point yields no additional improvement, as it no longer alleviates the long-tail bottleneck.

Building on the highest-throughput configuration (24Train16Infer) in \autoref{fig:Async_Async_Efficient}, we conduct ablation studies to analyze how individual components influence the optimal Async Ratio. As summarized in \autoref{tab:async_ratio}, the optimal Async Ratio is largely insensitive to model size, increases monotonically with sequence length, and decreases monotonically with rollout batch size. Surprisingly, a value as low as 2 suffices for most practical scenarios. \textbf{Interestingly, we can achieve substantial speedups from the Async framework without incurring significant off-policy penalties.}

\begin{tcolorbox}[colback=cyan!5!white, colframe=cyan!45!blue!60, title=\textbf{Takeaway 4: Async Training Can Be Stable and Nearly Performance-Lossless.}]
Under Async Ratio 2 and 8 settings, various off-policy methods, as well as widely used GRPO algorithm, can consistently deliver performance gains on par with synchronous training.
\end{tcolorbox}

While a small Async Ratio suffices under balanced workloads (Takeaway 3), a critical question remains: does increasing the Async Ratio hurt stability or final performance? We investigate this on \texttt{Qwen3-8B-Base} using standard GRPO-style training with small rollout batch size 32, evaluating popular off-policy algorithms under varying Async Ratios in a controlled setting.

\begin{figure}[t!]
    \centering

    % ============= 第一行：单个宽图（无 subcaption） =============
    \includegraphics[width=\textwidth]{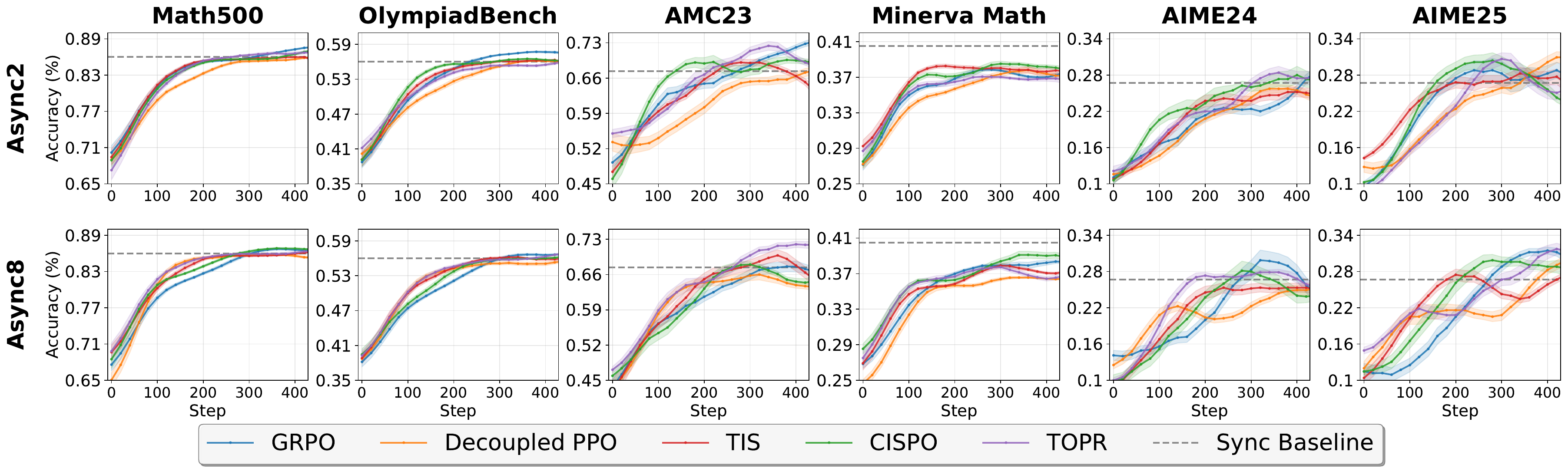}
    
    % \vspace{-1mm} % 增加一点垂直间距，避免拥挤

    % ============= 第二行：左右两个子图组 =============
    % \includegraphics[width=0.9\textwidth]{figures/RLVR/Async_Entropy_ResponseLength.pdf}
    % \vspace{-3mm}
    \caption{\textbf{Off-Policy Algorithm Performance Comparison under Async Ratio 2 and 8.} To ensure
    clarity and intuitiveness in the qualitative analysis, all curves are consistently smoothed using identical
    parameters. Specifically, the mean values are computed using an 11-step moving window. The shaded regions around the curves represent the range mean$\pm$ (std\_multiplier \(\times\) standard deviation), providing a visual representation of the oscillation amplitude. The Sync baseline uses the performance at 400 steps.} 
    \label{fig:off_policy_algorithm_performance}
\end{figure}

As shown in \autoref{fig:off_policy_algorithm_performance}, \textbf{all methods achieve comparable Pass@1 accuracy across benchmarks} and differences are minimal. Async variants slightly outperform the Sync baseline on \texttt{Math500} and \texttt{OlympiadBench}, while lagging marginally on \texttt{Minerva Math}. Notably, \textbf{vanilla GRPO alone yields strong performance}. Simply clipping tokens outside the target response region after importance sampling already provides a robust baseline. We also introduce \texttt{Weighted TOPR}, which improves stability by flexibly balancing positive and negative samples, thereby enhancing stability across diverse training scenarios.

In summary, \textbf{Async training reliably achieves competitive performance without relying on algorithm-specific tricks or heavy engineering}, demonstrating high throughput and training fidelity can coexist.

\section{Framework Design}
\label{sec:framework}

\subsection{Design Principles}
To fully harness the benefits of asynchronous training and provide users with a flexible programming model for asynchrony, we introduce \SysName{}, underpinned by two design principles: \emph{rollout--train decoupling} and \emph{fine-grained parallelism}. 

\noindent\textbf{Rollout-Train Decoupling.} Enabling asynchronous training requires managing staleness to prevent significant accuracy loss and allocating resources between rollout and training to maximize efficiency. To provide flexible control, we adopt a rollout--training decoupling architecture: execution workers for the two stages are placed on user-specified resources and run as a pipeline. At its core, users can configure the rollout model-update policy, transitioning from blocking, synchronous updates to non-blocking, asynchronous updates. Once updates become non-blocking, the rollout and training stages proceed in parallel, maximizing resource utilization and end-to-end throughput. Users can also adjust the frequency of asynchrony (e.g., asynchrony ratio) to mitigate accuracy loss.

\noindent\textbf{Fine-grained Parallelism.} We enable a fine-grained parallelism to execute LLM generation, environment interaction, and reward computation within the rollout stage. Instead of proceeding these phases in a full batch, fine-grained parallelism operates at the sample level. This allows users to control the lifecycle of each sample, determining when and where to execute each phase for a given sample. This enables a rollout pipeline where LLM generation for one sample overlaps with environment interaction for another and reward computation for a third. In addition, fine-grained parallelism distributes LLM generation workload evenly across GPUs via prompt replication, preventing long-tail rollouts from concentrating on a few devices and amplifying their adverse effects.

\begin{figure}[h!]
    \centering
    \includegraphics[width=0.95\textwidth]{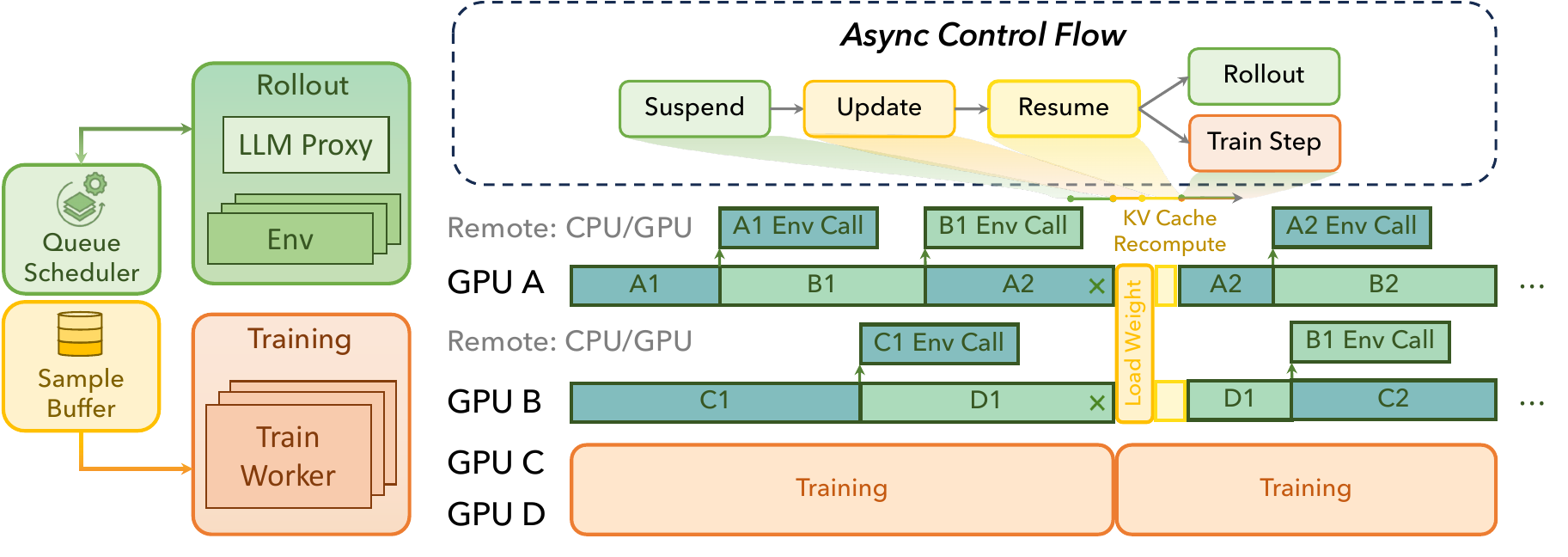}
    % \vspace{-10pt}
    % \vspace{-3mm}
    \caption{\textbf{Asynchronous Execution Workflow of \SysName{} for RLVR and Agentic Post-Training.} It consists of \green{\texttt{LLMProxy}}, \blue{\texttt{EnvManager}s}, \texttt{SampleBuffer}, and \purple{\texttt{AsyncController}}, which together orchestrate an asynchronous training workflow with fine-grained parallelism.}
    % \vspace{-3mm}
    \label{fig:pipeline}
\end{figure}
\subsection{Asynchronous Execution Workflow}

\autoref{fig:pipeline} illustrates the asynchronous execution workflow of \SysName{} for RLVR and agentic post-training. The asynchronous workflow centers on the rollout stage. Within the stage, the fine-grained parallelism maximizes the overlap between LLM generation, environment, and reward. Across stages, the rollout-train decoupling architecture parallelizes the execution of rollout and training. For clarity, we describe the system using the agentic RL training workflow.

\noindent\textbf{\green{\texttt{LLMProxy.}}} To orchestrate LLM inference, \SysName{} introduces the \green{\texttt{LLMProxy}}, which acts as an orchestrator for a fleet of internal backend workers and is shared by multiple \blue{\texttt{EnvManagers}}. Each worker centers around a command-driven event loop that manages an inference engine (e.g., vLLM). The loop is designed to maximize GPU utilization and enable full asynchrony. It operates continuously and non-blockingly, with three core services: (1) \textit{Step-wise Inference:} at each iteration, it advances the engine by executing a single decoding or prefill step over a batch of requests, saturating GPU resources. (2) \textit{Post-Processing:} whenever the engine completes a request, it immediately triggers a registered callback that post-processes the output and returns the result to the originating client (e.g., \blue{\texttt{EnvManager}}). (3) \textit{Process Commands:} The loop continuously proceeds commands dispatched from the proxy, \texttt{ADD} to enqueue new requests and \texttt{ABORT} to interrupt running requests and reclaim them into the \texttt{SampleBuffer} for subsequent recomputation and generation.

\noindent\textbf{\blue{\texttt{EnvManager.}}} It is the basic execution worker, enabling fine-grained parallel rollouts. Each \blue{\texttt{EnvManager}} starts a loop by resetting its environment via \texttt{reset}, then enters an independent event loop that mediates between its \texttt{BaseEnv} and the shared \green{\texttt{LLMProxy}}. In this loop, the \blue{\texttt{EnvManager}} receives the response as an action from the \green{\texttt{LLMProxy}}, applies it to \texttt{BaseEnv} via \texttt{step}, processes the resulting observation, and repeats until a termination condition is met.

With this fine-grained rollout, \SysName{} overlaps LLM decoding with the execution of thousands of environments. Upon trajectory completion, the \blue{\texttt{EnvManager}} immediately triggers reward computation, which proceeds in parallel with ongoing rollouts. By decoupling sample-level and environment-level execution, the design enables sample-level execution across components, achieving a high degree of parallelism and maximizing throughput.

\noindent\textbf{\purple{\texttt{AsyncController.}}} \SysName{} runs an asynchronous training pipeline via a \purple{\texttt{AsyncController}} and a shared \texttt{SampleBuffer}. A pool of \blue{\texttt{EnvManager}} processes act as independent producers: they generate trajectories and enqueue them into \texttt{SampleBuffer}. At each training step, the \purple{\texttt{AsyncController}} performs weight synchronization between the rollout and training stage in three phases: it issues \texttt{suspend} to pause trajectory collection, executes \texttt{model\_update} by fetching and broadcasting the latest weights to all LLM serving workers, and then sends \texttt{resume} so the \blue{\texttt{EnvManagers}} continue collecting trajectories with the updated model. In practice, the overhead of model update is a small fraction of total training time and does not impede rollout progress.

During each training iteration, the \purple{\texttt{AsyncController}} issues a blocking \texttt{get\_batch} to \texttt{SampleBuffer} to obtain a minibatch of trajectories, then executes \texttt{train\_step} on the retrieved data. In the  asynchronous mode, the training stage overlaps with the rollout stage, and the \blue{\texttt{EnvManagers}} together with the LLM serving workers continue collecting the next batch in parallel. 
\SysName{} can also be easily switched to synchronous mode: invoking \texttt{suspend} immediately after \texttt{get\_batch} pauses trajectory collection, ensuring that all subsequent trajectories are generated using the most up-to-date model weights.
Through this asynchronous design, users need not implement complex concurrency control or bespoke communication schemes. Optional barriers can be placed in \green{\texttt{LLMProxy}}, \blue{\texttt{EnvManager}}, and \purple{\texttt{AsyncController}} to support diverse training regimes (e.g., asynchronous training, batch rollout). In the absence of such barriers, the pipeline remains fully asynchronous, allowing the training process to continuously saturate available resources. Based on these components, we can configure an asynchronous ratio to control the degree of asynchrony, thereby achieving a trade-off between performance and training efficiency.

\subsection{Asynchronous Ratio}
\label{sec:async_ratio}

In \autoref{fig:overview} and \autoref{fig:pipeline}, we present our rollout--train decoupling architecture. In the \texttt{SampleBuffer}, response generation may be interrupted and resumed under newer policy LLMs. Consequently, response samples are generated from multiple policy LLM versions. Samples produced by stale policies can introduce high variance, undermining training stability. AReaL~\citep{fu2025areal} mitigates this by controlling the average sample freshness within a batch. In contrast, \SysName{} introduces \textbf{asynchronous ratio} $\alpha$ to regulate per-sample freshness. Specifically, \textbf{asynchronous ratio $\alpha$ is defined on per sample as the maximum allowable gap in policy version numbers} between the current policy and the policy version that initiated generation of that sample. If the policy network has advanced to version $n$, then any sample in \texttt{SampleBuffer} must have been initiated by a policy version no older than $(n - \alpha)$. Consequently, the \texttt{SampleBuffer} is upper-bounded by $(1+\alpha)\times\text{batchsize}$ samples and no sample is wasted, since we never generate samples that violate the freshness constraint. $\alpha$ is be a non-negative integer or real number.
% not necessarily an integer.

% \section{Features in \SysName{}}
\section{Detailed Design in RLVR and Agentic Pipeline}
\label{sec:key-feature}

\subsection{RLVR Pipeline}
\label{sec:RLVR}

\subsubsection{Queue Scheduling}
\label{sec:Queue_Scheduling}

In conventional RL post-training pipelines, rollouts are strictly synchronous and batched: a set of prompts is processed as one batch, and the LLM must complete generation for all prompts before any reward computation or filtering begins. This creates a straggler bottleneck because the longest sequence gates the batch, causing significant GPU underutilization, high rollout latency, and substantial overhead.

\begin{figure}[h]
    \centering
    \includegraphics[width=\textwidth]{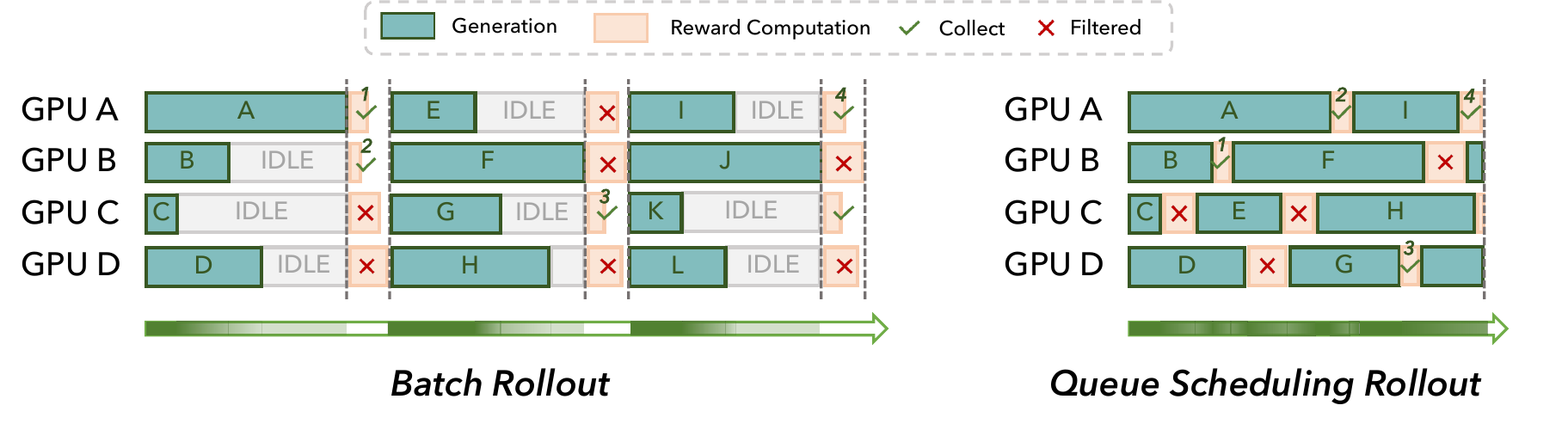}
    \vspace{-10pt}
    % \vspace{-3mm}
    \caption{\textbf{Comparison of Batch Rollout and Queue Scheduling Rollout.} Batch Rollout introduces substantial GPU idle time and leads to wasted generations when filtering is applied. In contrast, Queue Scheduling mitigates these issues by maintaining high GPU utilization throughout, computing rewards promptly, and terminating generation as soon as the desired number of qualifying samples is obtained.}
    % \vspace{-10pt}
    \label{fig:Queue_Scheduling}
\end{figure}

\SysName{} focuses on fine-grained parallelism and employs \texttt{Queue Scheduling} to address these limitations. Each prompt is treated as an independent rollout task and enqueued for dynamic scheduling. Once a response is generated, it is immediately dispatched to a reward worker for evaluation, without waiting for the remainder of the batch. Reward computation overlaps with ongoing generation, which removes pipeline bubbles and reduces GPU idle time. This design delivers two key benefits:
(1) it dramatically improves GPU utilization by keeping compute resources continuously engaged across responses with various lengths;
(2) in dynamic filtering scenarios with redundant prompts, it accelerates the collection of high-quality samples, thereby increasing overall training throughput. \autoref{fig:Queue_Scheduling} clearly illustrates the advantages conferred by queue scheduling rollout.

\begin{figure}[h!]
    \centering
    \includegraphics[width=0.95\textwidth]{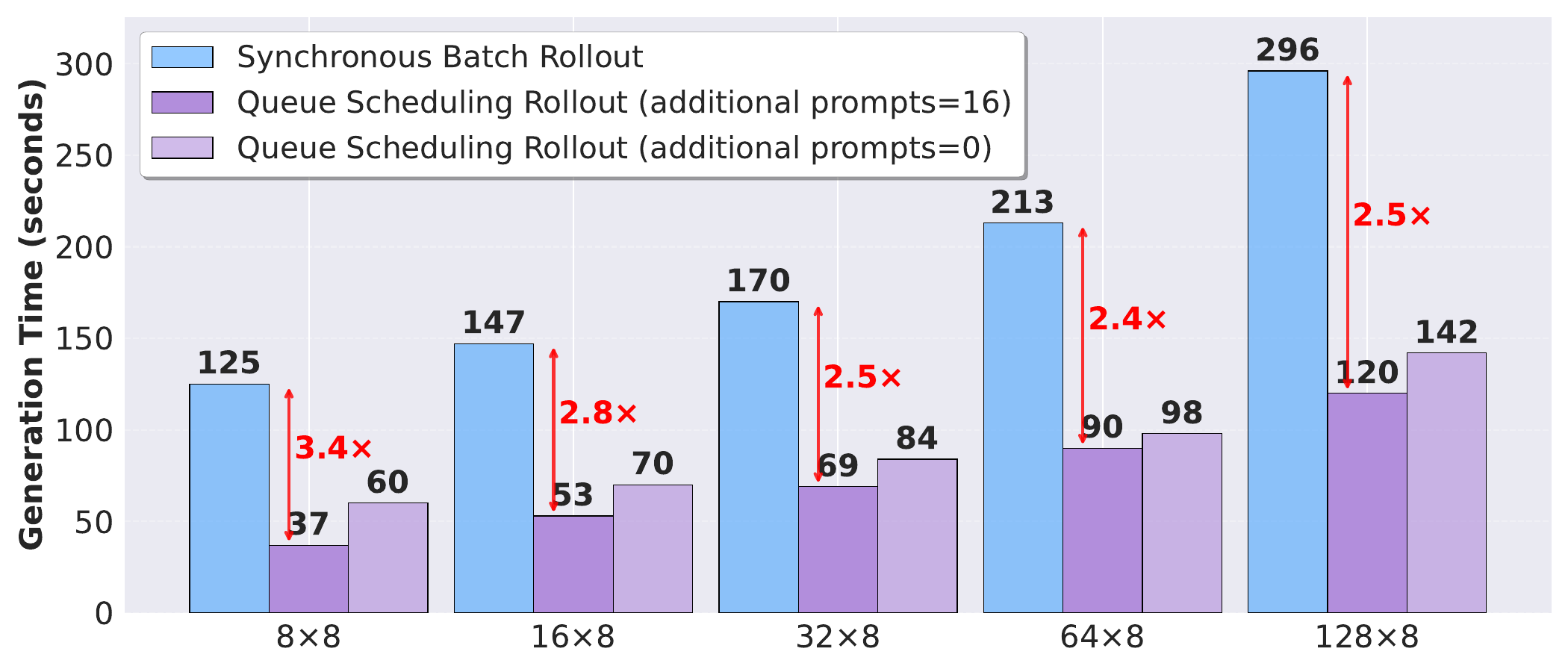}
    \vspace{-5pt}
    % \vspace{-3mm}
    \caption{\textbf{Efficiency comparison of generation time across different batch size$\times8$ configurations.} The blue bars represent conventional \texttt{Synchronous Batch Rollout}, while the purple bars show \texttt{Queue Scheduling Rollout} with \texttt{max\_additional\_running\_prompts} set as $16$ and $0$,
    respectively. Red double-headed arrows indicate speedup ratios using  \texttt{Queue Scheduling Rollout} (\texttt{additional prompts}$=16$).}
    \vspace{-3mm}
    \label{fig:queue_exp}
\end{figure}

\paragraph{Experimental Evaluation.} 
We empirically evaluate the effectiveness of queue scheduling under dynamic filtering. In the synchronous baseline, reward computation is deferred until the entire batch completes generation. In our setup, we generate \(k = 8\) responses per prompt, allow up to 16 additional concurrent prompts, and filter out samples with zero intra-group variance. We compare queue scheduling, with and without redundant generation, against the baseline across varying batch sizes. As shown in \autoref{fig:queue_exp}, queue scheduling reduces average per-step generation time. For example, with 16 redundant prompts and an \(8 \times 8\) configuration (8 prompts, each with 8 responses), the average per-step generation time drops from 125 seconds to 37 seconds (3.4× speedup). Similar gains are observed for larger batch sizes, and the benefit grows with higher redundancy and stronger filtering. These results confirm that queue scheduling effectively improves rollout pipeline efficiency, especially in dynamic filtering scenarios.

\subsubsection{Prompt Replication}
\label{sec:prompt_replication}

\SysName{} implements prompt replication to further improve rollout efficiency. This mechanism alleviates the synchronization bottlenecks inherent in multi-candidate decoding. Prior work~\citep{hybridflow} typically sets \texttt{num\_return\_sequences} \(>\) 1 to generate multiple responses for a single prompt during rollout~\citep{shao2024deepseekmath}, which forces a single worker to synchronously decode all \(n\) responses. \SysName{} instead expands each prompt into \(n\) independent rollout tasks, each producing a single response, via the flag \texttt{is\_num\_return\_sequences\_expand}. This decoupling allows candidates from the same prompt to run on separate GPUs and be scheduled independently, reducing pipeline bubbles caused by heterogeneous response lengths. As illustrated in \autoref{fig:overview}, replicating prompt C and scheduling its candidate responses (C1 and C2) on different GPUs effectively reduces these bubbles.

\paragraph{Experimental Evaluation.} 

\begin{figure}[h!]
    \centering
    \includegraphics[width=1\textwidth]{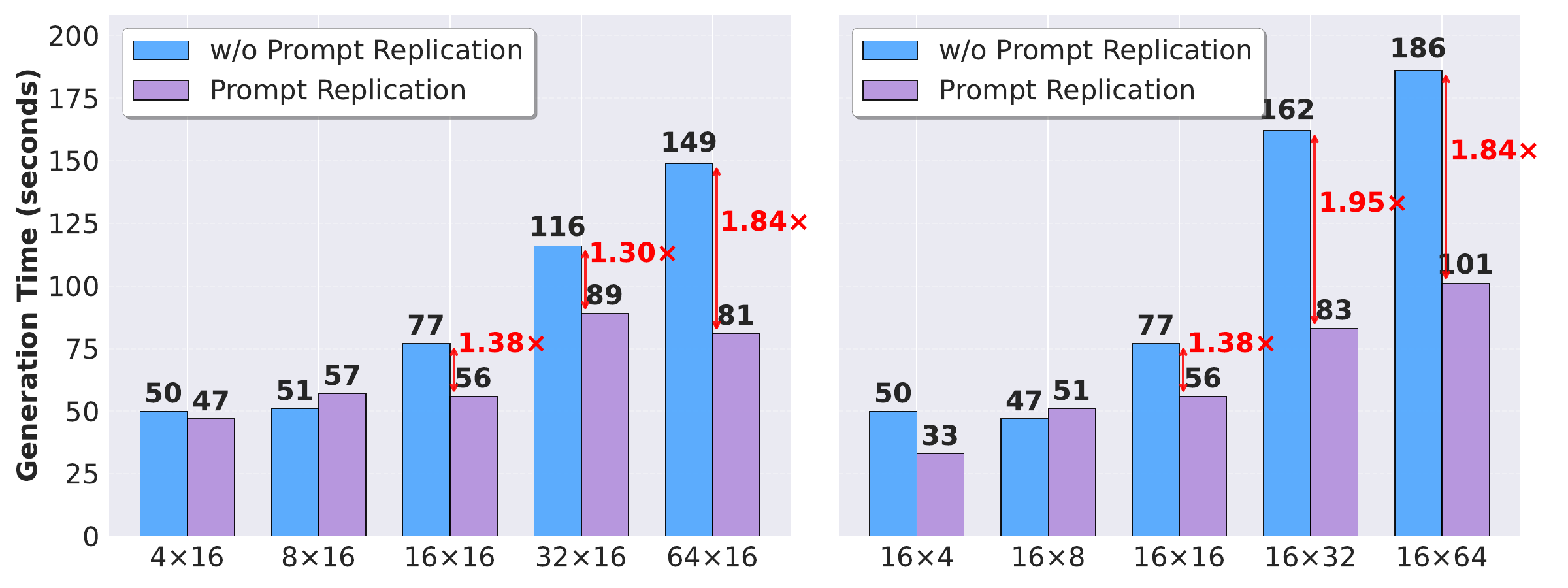}
    \caption{\textbf{Efficiency of using prompt replication across different rollout configurations.} \textbf{Left:} Varying \texttt{batch size} with \texttt{num\_return\_sequences}=16. \textbf{Right:} Varying \texttt{num\_return\_sequences} with \texttt{batch size}=16. In both cases, prompt replication substantially reduces generation time by alleviating straggler effects, achieving up to $1.84\times$ speedup under large-batch or large-response per prompt configurations.
    }
    \vspace{-1mm}
    \label{fig:prompt_replication}
\end{figure}

We quantify the impact of prompt replication under varying \texttt{batch\_size} and \texttt{num\_return\_sequences} configurations. Firstly, we fix \texttt{num\_return\_sequences} as 16 and scale \texttt{batch\_size} from 4 to 64. As shown in \autoref{fig:prompt_replication}, prompt replication yields limited gains at small batch sizes but delivers substantial improvements beyond moderate scales by mitigating long-tail stragglers and reducing mean step time. For instance, at $32 \times 16$, latency drops from 116 seconds to 89 seconds ($1.30\times$ speedup); at $64 \times 16$, latency reduces from 149 seconds to 81 seconds ($1.84\times$ speedup). 
Secondly, we fix \texttt{batch\_size} as 16 and increase \texttt{num\_return\_sequences} from 4 to 64. Prompt replication consistently enhances efficiency as the number of candidates grows. At $16 \times 32$, step time decreases from 162 s to 83 s (1.95\(\times\) speedup), and at $16 \times 64$, it still achieves a $1.84\times$ speedup. Overall, 
these results confirm that \textbf{prompt replication enables fine-grained intra-rollout parallelism, effectively delivering significant efficiency gains.}

\subsection{Agentic Pipeline}
\label{sec:agentic}

In agentic pipelines, a single trajectory involves multiple rounds of interaction with complex external environments, such as \textit{SWE}~\citep{jimenez2023swe}, \textit{ALFWorld}~\citep{shridhar2020alfworld}, and \textit{ShopSimulator}~\citep{wang2025shopsimulator}, where execution latency varies widely and failures are common. Although most rollouts complete within seconds, some extend to minutes due to environment initialization and network latency. This pronounced long-tail latency significantly degrades the training efficiency and motivates two key designs: \emph{environment-level asynchronous rollout} and \emph{redundant environment rollout}.

\subsubsection{Environment-Level Asynchronous Rollout}
\label{sec:agentic_async}
To reduce GPU idleness during environment interactions, we devise an \emph{environment-level asynchronous rollout}. We decompose each trajectory into a sequence of fine-grained, environment-level interaction units. Once a trajectory begins interacting with an environment to receive feedback, the pending trajectories in the \texttt{SampleBuffer} are immediately dispatched to available LLM serving workers for continued response (i.e., action) generation.

\paragraph{Experimental Evaluation.} 

\begin{wrapfigure}{r}{0.5\textwidth} \centering \includegraphics[width=0.5\textwidth]{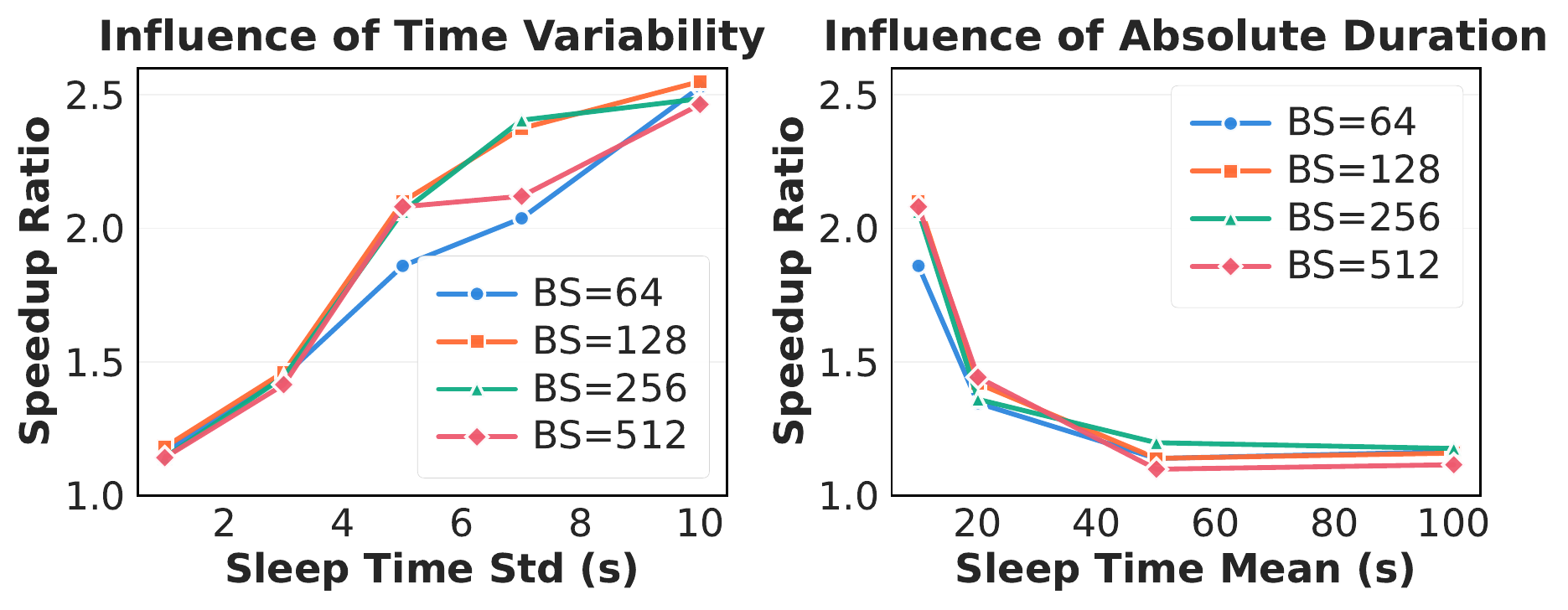} 
\vspace{-15pt}
\caption{\textbf{Simulation results of environment-level asynchronous rollout under varying environment latency distributions.} 
\textbf{Left:} Speedup increases with higher standard deviation of environment latency at fixed mean ($\mu=10$s). 
\textbf{Right:} Speedup decreases as the mean step time grows while keeping std fixed ($\sigma=5$s), since the impact of stragglers diminishes.} \vspace{-5mm} \label{fig:agentic_async_train} \end{wrapfigure}

We first conduct controlled simulations in which environment latencies are sampled from Gaussian distributions with mean $(\mu)$ and standard deviation $(\sigma)$. As shown in \autoref{fig:agentic_async_train}, a clear trend emerges: \textbf{larger variance yields greater speedup}. When latencies are nearly uniform, e.g., $(10,1)$, the benefit is modest, limited to $1.16\times$ at batch size~512. As variance increases, the gains become substantial. With $(10,10)$, the average step time drops from $892$ seconds to $362$ seconds at batch size~512, a $2.46\times$ improvement. Similar results hold for $(10,7)$, where the speedup reaches $2.12\times$, and for $(50,5)$, where it is $1.20\times$. These results show that asynchronous scheduling shortens overall step time and sustains high throughput, with benefits that grow as environment latency variance increases.

We further validate this mechanism in real environments. As shown in \autoref{fig:env_rep_real}, even in Sync training, environment-level asynchronous rollout can reduce the end-to-end training time from $10.22$h to $8.32$h on SWE ($1.23\times$) and from $13.37$h to $8.44$h on ALFWorld ($1.58\times$). These results confirm that \textbf{environment-level asynchronous rollout is consistently effective beyond simulation and brings satisfactory gains in practice.} Detailed experimental configurations are provided in \autoref{app:training_details}.

\subsubsection{Redundant Environment Rollout}
\label{sec:agentic_envrep}

We introduce \emph{redundant environment rollout} to mitigate the negative impact of environment instability on agentic RL training efficiency. This mechanism offers two tunable controls: (1) increasing \texttt{num\_env\_groups} to spawn more concurrent environment groups, and (2) increasing \texttt{group\_size} to generate more candidate trajectories per group. Since \SysName{} terminates rollout once a predefined number of trajectories has been collected, increasing \texttt{num\_env\_groups} and \texttt{group\_size} helps prevent fail-slow and fail-stop environments from becoming system bottlenecks. Empirically, we observe that increasing \texttt{num\_env\_groups} can deliver stronger resilience to fail-slow and fail-stop behavior than increasing \texttt{group\_size}.

\paragraph{Experimental Evaluation.}

\begin{figure}[htbp]
    \centering
    \newlength{\figheight}
    \setlength{\figheight}{4.83cm}

    \begin{minipage}[t]{0.37\textwidth}
        \centering
        \adjustbox{height=\figheight, valign=t}{\includegraphics[width=\textwidth]{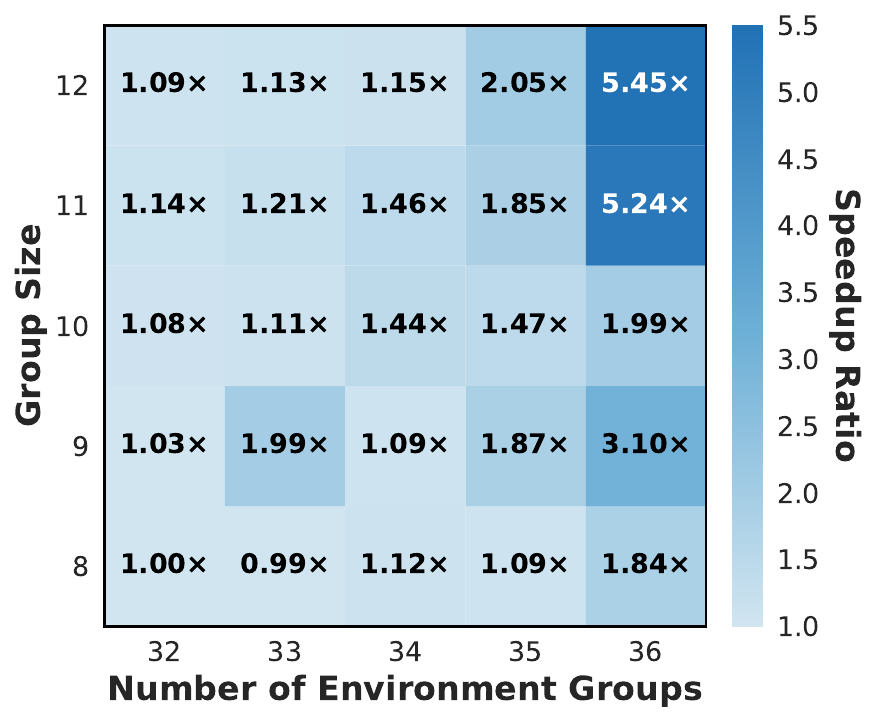}}\\
        % \captionsetup{justification=centering, font=bf}
        \caption{\textbf{Heatmap of speedup across group size and env group count.}}
        \label{fig:env_redundant}
    \end{minipage}
    \hfill
    \begin{minipage}[t]{0.61\textwidth}
        \centering
        \adjustbox{height=\figheight, valign=t}{\includegraphics[width=\textwidth]{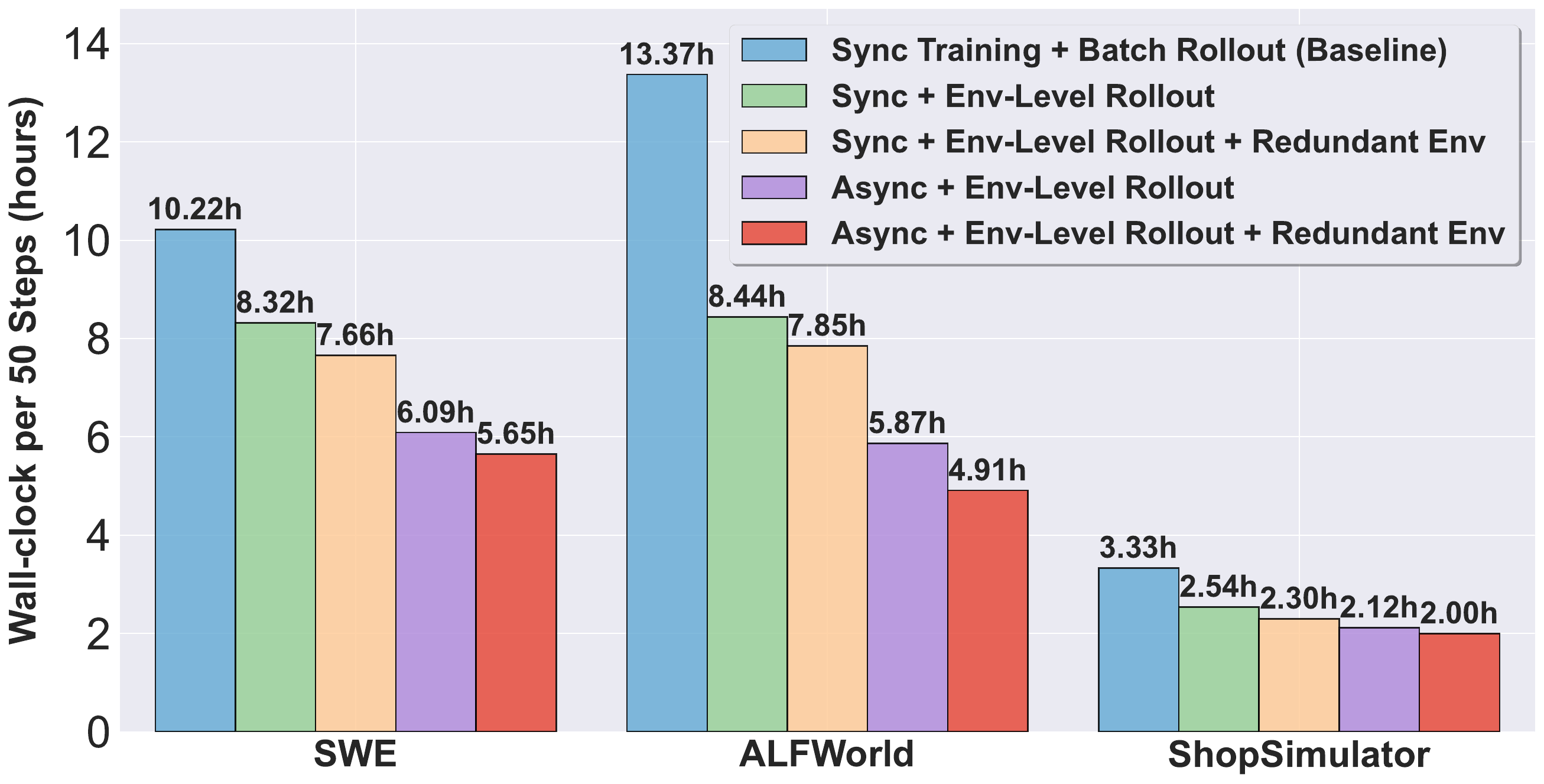}}\\
        % \captionsetup{justification=centering, font=bf}
        \caption{\textbf{Real-environment evaluation of environment-level asynchronous and redundant environment rollout.}}
        \label{fig:env_rep_real}
    \end{minipage}
\end{figure}

We simulate different configurations by fixing the total rollout batch size at $256$ and varying \texttt{num\_env\_groups} and \texttt{group\_size} where environment latency is modeled by Gaussian distributions with varying mean $\mu = 10$ and standard deviation $\sigma = 5$. The results in \autoref{fig:env_redundant} show that \textbf{increasing the number of groups is consistently more effective than enlarging group size.} For instance, scaling from $32\times 8$ (baseline) to $36\times 12$ reduces step time from $243$ seconds to $45$ seconds, a $5.45\times$ speedup. Similar improvements are observed for $36\times 11$ ($5.24\times$) and $36\times 9$ ($3.10\times$). The heatmap visualization highlights that higher group counts lead to more stable step times and better robustness against latency variance.

We also validate this design in real environments. As shown in \autoref{fig:env_rep_real}, \textbf{redundant environment rollout yields additional gains on top of both synchronous and asynchronous rollouts.} On SWE, the training time reduces from $8.32$h to $7.66$h under synchronous rollout ($-7.9\%$), and from $6.09$h to $5.65$h under asynchronous rollouts ($-7.2\%$). On ALFWorld, the corresponding reductions are from $8.44$h to $7.85$h ($-7.0\%$) and from $5.87$h to $4.91$h ($-16.4\%$), respectively. \textbf{These results demonstrate that redundant environment rollout complements env-level asynchronous rollout}, providing an extra $7\%$-$16\%$ throughput improvement in real agentic environments. Together, both techniques form an effective design to sustain training efficiency under stochastic and failure-prone conditions..

% Full configuration details are provided in \autoref{app:training_details}.

% \input{content/6_discussion}
\section{Conclusion}

In this report, we present theoretical and empirical evidence for the benefits of asynchronous training, motivating the design of \SysName{}, which extends ROLL with native support for asynchrony. \SysName{} is grounded in two core design principles: \emph{fine-grained parallelism} and \emph{decoupling between rollout and training}. Guided by these principles, \SysName{} implements queue scheduling, prompt replication, and an asynchronous training architecture. In particular, \SysName{} introduces environment-level asynchronous rollout and redundant environment rollout to expedite the agentic RL pipeline. Our extensive experiments demonstrate the efficiency of \SysName{}.

% \begin{figure}[h]
%     \centering

%     % ============= 第一行：单个宽图（无 subcaption） =============
%     \includegraphics[width=0.95\textwidth]{figures/gen.pdf}
    
%     % \vspace{-1mm} % 增加一点垂直间距，避免拥挤

%     % ============= 第二行：左右两个子图组 =============
%     % \includegraphics[width=0.9\textwidth]{figures/RLVR/Async_Entropy_ResponseLength.pdf}
%     % \vspace{-3mm}
%     \caption{\textbf{An illustration of training acceleration with ROLL Flash.}}
%     \label{fig:gen_images}
% \end{figure}
% \input{content/authors.tex}

\clearpage
\bibliography{biblio}
\bibliographystyle{colm2024_conference}

\clearpage
\appendix
\renewcommand{\sectionautorefname}{Appendix}
\renewcommand{\subsectionautorefname}{Appendix}

\section{Training Details}
\label{app:training_details}

\subsection{RLVR Pipeline}
\paragraph{Datasets.}
We use \textsc{DAPO-Math-18K}~\citep{yu2025dapo} as the training dataset. For evaluation, we use \textsc{MATH-500}, \textsc{OlympiadBench}, \textsc{MinervaMath}, \textsc{AMC 2023}, \textsc{AIME 2024}, and \textsc{AIME 2025}.

\paragraph{Implementation Details.}  The \texttt{async\_generation\_ratio} controls asynchrony: 0 denotes synchronous (Sync) mode, while any positive integer or fractional value specifies the async ratio. In async mode, GPU allocations for generation and training are set via \texttt{actor\_infer} and \texttt{actor\_train}, respectively. Advantage estimates are computed using group-normalized rewards and applied across off-policy algorithms selected by \texttt{pg\_variant}.  We use \textsc{SGLang}~\citep{zheng2024sglang} v0.4.6 and \textsc{vLLM}~\citep{kwon2023efficient} v0.8.4 as generation backends, and Megatron~\citep{shoeybi2019megatron} for distributed training. 

\begin{codeblock}
\begin{verbatim} 
seed: 42
pg_variant: ppo  # can be decoupled_ppo, topr, tis, cispo
gamma: 1.0        # discount factor
lambd: 1.0        # GAE lambda

pretrain: Qwen/Qwen3-8B-Base  

rollout_batch_size: 256               # prompt count
num_return_sequences_in_group: 16  # group size per prompt
ppo_epochs: 1                            # per sample usage

prompt_length: 2048
response_length: 30720

generate_opt_level: 0                       # whether to use Queue Scheduling
is_num_return_sequences_expand: false   # whether to use Prompt Replication

async_generation_ratio: 0                  # 0 represnets Sync, > 0 represnet Async

# use GRPO
adv_estimator: "reinforce"
reward_norm: group

actor_train:
    data_args:
        template: qwen2_5
        file_name:
        - data/train_data_math_dapo_18k.jsonl  # use DAPO dataset
    training_args:
        learning_rate: 1.0e-6
        weight_decay: 0
        per_device_train_batch_size: 1
        gradient_accumulation_steps: 256  # control on-policy or 4 minibatchs update
        warmup_steps: 20
        # Use Train Speed Up
        use_remove_padding: true
        use_dynamic_batching_in_train: true
    device_mapping: list(range(0,16))    
actor_infer:
    generating_args:
        max_new_tokens: ${response_length}
        top_p: 1
        top_k: 1000000
        num_beams: 1
        temperature: 1
        num_return_sequences: ${num_return_sequences_in_group}
    device_mapping: list(range(0,16))   # can be different from train in Async Setting
...
\end{verbatim} 
\end{codeblock}

In practice, we enforce temperature~=~1 and top-$p$~=~$1$ to obtain raw, unmodified logits, following the same practice as AReaL~\citep{fu2025areal}. This is necessary because our inference engine must produce the original token probabilities; any modification of sampling parameters would alter the output distribution and prevent us from recovering the true logits. While this ensures fidelity, it also restricts the flexibility of sampling hyperparameters. The limitation can be resolved with future architectural interface support.

Moreover, due to inherent discrepancies between the inference engine (e.g., vLLM or SGLang) and the training engine (e.g., Megatron), we adopt truncated importance sampling (IS) to stabilize training. Specifically, we cap the importance weight at a threshold $C$ (e.g., $C = 5$):
\begin{equation}
\min\left( \frac{\textcolor{blue}{\pi_{\text{megatron}}}(a \mid \theta)}{\textcolor{red}{\pi_{\text{vllm}}}(a \mid \theta)},\, C \right).
\end{equation}
This issue also arises in synchronous (Sync) architectures, and we address it using the same off-policy correction as in VeRL~\citep{yao2025offpolicy}.

\subsection{Agentic Pipeline}
\label{app:config_agentic}
\paragraph{Datasets.} We utilized R2E-Gym-Lite \citep{jain2025r2egymproceduralenvironmentshybrid} as the training dataset for the SWE domain. For the others, we employed ALFWorld-Train \citep{shridhar2020alfworld} and ShopSimulator-SingleTurn \citep{wang2025shopsimulator} as the training datasets for ALFWorld and ShopSimulator, respectively. During evaluation, we adopted SWE-Bench-Verified \citep{jimenez2023swe}, ALFWorld, and ShopSimulator-SingleTurn as the test benchmarks to ensure comprehensive and consistent assessment across distinct task domains.

\paragraph{Implementation Details.} The async ratio and generation backend adopt the same implementation details as employed in the RLVR Pipeline. Beyond these shared components, we introduce a Redundant Env experimental configuration. In the default mode, the \texttt{train\_env\_manager} satisfies the relation \texttt{group\_size}~\(\times\)~\texttt{num\_env\_groups} = \texttt{rollout\_batch\_size}, whereas in the Redundant Env mode this condition is relaxed such that \texttt{group\_size}~\(\times\)~\texttt{num\_env\_groups} \(>\) \texttt{rollout\_batch\_size}. Concretely, across all three scenarios, the default configuration sets the \texttt{train\_env\_manager} with \texttt{group\_size = 16} and \texttt{num\_env\_groups = 8}, and the \texttt{val\_env\_manager} with \texttt{group\_size = 1} and \texttt{num\_env\_groups = 128}. Under the Redundant Env mode, the settings are adjusted to \texttt{train\_env\_manager: group\_size = 17, num\_env\_groups = 9} and \texttt{val\_env\_manager: group\_size = 1, num\_env\_groups = 144}.

\begin{codeblock}
\begin{verbatim} 
pretrain: Qwen/Qwen3-8B         #  Qwen/Qwen3-14B for SWE, Qwen/Qwen3-8B for others
async_generation_ratio: 1       #  0 represnets Sync, > 0 represnet Async

rollout_batch_size: 128         # env count for train
sequence_length: 32768          # max sequence length

train_env_manager:
  num_env_groups: 8              # can be different from train in Redundant Env
  group_size: 16                  # can be different from train in Redundant Env
val_env_manager:
  num_env_groups: 128            # > ${val_batch_size} for Redundant Env
  group_size: 1

actor_train:
  device_mapping: list(range(0,32))      
actor_infer:
  device_mapping: list(range(32,64))     

# Env Special Parameters
custom_envs:
  SWEEnv:
    max_steps: 50
    max_new_tokens: 8192
  AlfworldEnv:
    max_steps: 30
    max_new_tokens: 4096
  ShopSimulator:
    max_steps: 30
    max_new_tokens: 2048
...
\end{verbatim} 
\end{codeblock}

\end{document}